\documentclass[10pt,journal,twocolumn]{IEEEtran}
\usepackage{graphicx}
\usepackage{amstext}
\usepackage{amsmath}
\usepackage{amsfonts,amssymb}
\usepackage{amsmath,tabu}
\usepackage{amssymb}
\usepackage[dvips]{epsfig}
\usepackage{epsfig}
\usepackage[dvips]{graphicx}
\usepackage{multirow}
\usepackage{multirow} 
\usepackage[utf8]{inputenc}
\usepackage[english]{babel}
\usepackage{caption}
\usepackage{subcaption}
\usepackage{float}
\usepackage{ltablex}
\usepackage{algorithmic}
\usepackage{setspace}
\usepackage[compact]{titlesec}
\titlespacing{\subsection}{0pt}{1ex}{1ex}
\newtheorem{theorem}{Theorem}[section]

\newtheorem{lemma}[theorem]{Lemma}

\def\b1{{\boldsymbol{1}}}
\def\c1{{\textcircled{a}}}
\def\ba{{\boldsymbol{a}}}

\def\bc{{\boldsymbol{c}}}

\def\be{{\boldsymbol{e}}}
\def\boldf{{\boldsymbol{f}}}

\def\bh{{\boldsymbol{h}}}

\def\bq{{\boldsymbol{q}}}

\def\bs{{\boldsymbol{s}}}

\def\bv{{\boldsymbol{v}}}

\def\bx{{\boldsymbol{x}}}
\def\by{{\mathbf{y}}}
\def\bz{{\boldsymbol{z}}}
\def\bA{{\mathbf{A}}}
\def\bB{{\boldsymbol{B}}}

\def\bG{{\boldsymbol{G}}}

\def\bI{{\mathbf{I}}}

\def\bW{{\boldsymbol{W}}}

\providecommand{\keywords}[1]{\textbf{\textit{Index terms---}} #1}
\begin{document}
\title{PROMPT: Parallel Iterative Algorithm for $\ell_{p}$ norm linear regression via Majorization Minimization with an application to semi-supervised graph learning}
\author{R.~Jyothi and P.~Babu}
\maketitle
\begin{abstract}
In this paper, we consider the problem of $\ell_{p}$ norm linear regression, which has several applications such as in sparse recovery, data clustering, and semi-supervised learning. The problem, even though convex, does not enjoy a closed-form solution. The state-of-the-art algorithms are iterative but suffer from convergence issues, i.e., they either diverge for $p>3$ or the convergence to the optimal solution is sensitive to the initialization of the algorithm. Also, these algorithms are not generalizable to every possible value of $p$. In this paper, we propose an iterative algorithm : \textbf{P}arallel Ite\textbf{R}ative Alg\textbf{O}rith\textbf{M} for $\ell_{\textbf{P}}$ norm regression via Majoriza\textbf{T}ion Minimization (\textbf{PROMPT}) based on the principle of Majorization Minimization and prove that the proposed algorithm is monotonic and converges to the optimal solution of the problem for any value of $p$. The proposed algorithm can also parallelly update each element of the regression variable, which helps to handle large scale data efficiently, a common scenario in this era of data explosion. Subsequently, we show that the proposed algorithm can also be applied for the graph based semi-supervised learning problem. We show through numerical simulations that the proposed algorithm converges to the optimal solution for any random initialization and also performs better than the state-of-the-art algorithms in terms of speed of convergence. We also evaluate the performance of the proposed algorithm using simulated and real data for the graph based semi-supervised learning problem.
\end{abstract}
\keywords{$\ell_{p}$ norm linear regression, Majorization Minimization, Parallel algorithm, Graph based semi-superivsed learning}
\section{Introduction and Problem Formulation} \label{sec:1}
\begin{figure}[!h]
\centering
\begin{tabular}{c}
\includegraphics[height=1.5in,width=3.1in]{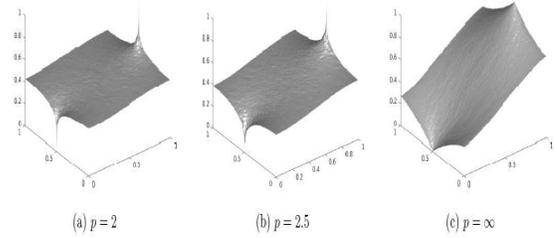}
\end{tabular}
\caption{Illustration of the function surface $\hat{f}(v)$ for different values of $p$. For $p=2$ the surface is constant with peaks near the labeled points and for $p>2$ the surface becomes smoother. This example was generated for two labeled and 105 unlabeled data points \cite{convergence}.\vspace{-2mm}}
\label{illustration}
\end{figure}

Linear regression is a statistical model which is used to estimate the relationship between a dependent variable and one or more independent variables \cite{bishop}. This technique is commonly used in data forecasting \cite{forecasting}, time series analysis \cite{timeseries} and for risk assessment \cite{risk}. In linear regression, the dependent variable $y$ is related to the unknown independent variables $\bx= [x_{1},x_{2}, \ldots, x_{n}]^{T} \in \mathbb{R}^{n}$ using a linear model as follows: 
\begin{equation}\label{model}
    \begin{array}{ll}
         y=a_{1}x_{1} + a_{2}x_{2}+ \cdots +a_{n}x_{n} + e = \ba^{T}\bx + e
    \end{array}
\end{equation}
where $a_{i}$'s are the known regression parameters and $e$ is the unknown noise usually modeled as a Gaussian random variable. Suppose we have $m$ such observations, then the model in (\ref{model}) can be re-written compactly as:
\begin{equation}\label{model-compact}
\begin{array}{ll}
\by = \bA\bx +\be
\end{array}
\end{equation}
 where $\by \in \mathbb{R}^{m}$ is obtained by stacking the observations $\{y_{i}\}_{i=1}^{m}$, $\bA = [\ba_{1}, \ba_{2}, \ldots, \ba_{m}]^{T} \in \mathbb{R}^{m \times n}$ and $\be=[e_{1}, e_{2}, \ldots, e_{m}]^{T} \in \mathbb{R}^{m}$ is the unknown error vector. 
 Usually, $m>n$ i.e., there are more equations than unknowns and hence the linear system of equations in (\ref{model-compact}) does not have any solution. In such a case, a classical approach adopted to solve the set of equations in (\ref{model-compact}) is to find an $\bx$ such that the sum of the squares of the error i.e., the $\ell_{2}$ norm of the error vector is minimum: 
\begin{equation} \label{eq:11}
\begin{array}{ll}
\underset{\bx}{\rm arg\: min} \: \{f(\bx)= \|\by-\bA\bx\|^{2}_{2}\}
\end{array}
\end{equation}
where $\|\cdot\|_{2}$ is the Euclidean norm. The estimate $\bx$ obtained by solving the above problem is known as the least-squares estimator and is given by $\bx = \left(\bA^{T}\bA\right)^{-1}\left(\bA^{T}\by\right)$. However, it has been observed that the performance of the least-squares estimator is degraded when the underlying noise is not Gaussian distributed \cite{lpnorm}. To overcome this issue, the authors in \cite{r1} and \cite{r2} show that the estimate $\bx$  obtained by minimizing the $\ell_{p}$ norm of the error vector, for $p \neq 2$, is less sensitive to the noise distribution. In practice, the noise distribution may be non-Gaussian and hence in this paper we focus on solving the $\ell_{p}$ norm linear regression problem which is defined as:
\begin{equation} \label{eq:12}
\begin{array}{ll}
\textrm{LR:} \quad \underset{\bx}{\rm arg\: min} \: \{f_{_{LR}}(\bx)= \|\by-\bA\bx\|_{p}\}
\end{array}
\end{equation}
Even though the problem in (\ref{eq:12}) is convex and differentiable, one cannot obtain a closed-form solution for it using the KKT conditions. Hence, the state-of-the art methods used to solve the problem in (\ref{eq:12}) are iterative in nature. The benchmarking algorithm used to solve the $\ell_{p}$ norm problem for $p \in [1, \infty)$ is the Iterative Reweighted Least Squares (IRLS) algorithm (\cite{irls4, irls1,irls2, irls5}). It is a simple iterative algorithm which replaces the objective function in (\ref{eq:12}) with a weighted $\ell_{2}$ norm function at every iteration, which when minimized admits a closed-form solution. However, it was found that this method diverges for $p>3$ (\cite{irls2},\cite{convergence}). Since then, many algorithms have been proposed to overcome the issue of convergence faced by IRLS algorithm (\cite{modifiedirls1, modifiedirls2, firls}). We review these methods briefly in the next section. Apart from being a robust regressor, the $\ell_{p}$ norm linear regression for $p>2$ has applications in  semi-supervised learning commonly used in classification problems (\cite{ssl,app, clustering}), which we discuss next.\\
In today's world, classification problems are becoming pervasive wherein the goal is to categorize the given data into its corresponding class using a mapping function \cite{bishop}. In the supervised learning setting, the training data set (which consists of data points and its corresponding class labels) are used to learn the mapping function which can be used later to predict the class label of an unknown test data point. However, one setback of this approach is that it requires large amount of labeled data points to learn the mapping function. Data labeling is a laborious task (e.g. writing a transcript for speech recognition) and also requires expert input (deciding whether a brain scan is healthy or not). However, one can easily obtain large amounts of unlabeled data. Semi-supervised learning utilizes both the labeled and unlabeled data points to learn the mapping function. By doing so, semi-supervised learning learns a better mapping function when compared to the supervised learning which uses only the labeled data points \cite{convergence}.\\
Under the semi-supervised learning approach, graph based semi-supervised learning is commonly used, wherein the labels of the unknown data points are learned by propagating the known label information using graphs. Suppose we are given $n$ data points $\bv_{i} \in \mathbb{R}^{d}$ out of which $u$ are unlabeled data points and $l$ are labeled points with labels $h_{i}=f(\bv_{i})$ (where $f$ is an unknown real-valued function). Generally, the number of labeled points are much smaller than the number of unlabeled points i.e., $l<<u$.  
An undirected graph $\bG = (V,E)$  with vertex set $V=\{v_{1},v_{2},\ldots v_{n}\}$ are used to represent the $n$ data points. The subset ${L}\subset V$ of the vertex set are used to represent the labeled points and the remaining vertices $V\setminus L$ represents the unlabeled points. The edges $E$ of the graph are equipped with non-negative edge weights ${W}=\{w_{xy}\}_{x,y \in {E}}$, which are chosen such that $w_{xy}\approx 1$ when $x$ is similar to $y$ and $w_{xy}\approx 0$ when $x$ and $y$ are dissimilar. The task of graph based semi-supervised learning is to assign labels to the remaining vertices without changing the labels at the vertex subset $L$. Since, this problem has infinitely many solutions, one usually makes the \emph{semi-supervised smoothness assumption} which states that if two data points $\bv_{1}$, $\bv_{2}$ in a high density region are close then so should be their corresponding outputs $h_{1}$, $h_{2}$; hence the function $f$ is required to be smooth in the high density regions \cite{sslbook}. A common approach used to label the data points at remaining vertices such that the mapping function does not change rapidly in the high density regions and also agrees with the labels at the vertices $L$ is by  using the $\ell_{2}$-based Laplacian regularizer, which is given as:
\begin{equation}\label{l2graph}
\begin{array}{ll}
\hat{f}(v) = \underset{f}{\rm arg\: min} \:\displaystyle\sum_{i,j \in E} w_{ij} \left(f(v_{i})-f(v_{j})\right)^{2}\\
\hspace{10mm} {\rm{subject\:to}}\: f(v_{i}) = h_{i} , i \in {L}
\end{array}
\end{equation}
The above model was first introduced by \cite{laplace}. From an intuitive perspective, minimizing the objective function in (\ref{l2graph}) will ensure that the
similar points (for which $w_{ij}$ is larger) will have similar labels, making $\left(f(v_{i})-f(v_{j})\right)^{2}$
smaller, while for dissimilar points ($w_{ij}$ is small) and hence are allowed to have dissimilar labels. However, it was found that when the number of unlabeled data points $u$ is much greater than the number of labeled data points $l$, the learned function $\hat{f}(v)$ becomes constant everywhere, with sharp spikes near the labeled data. We illustrate the same in Fig.\ref{illustration}.a.  for the number of labeled points $l$ equal to $2$ and number of unlabeled points $u$ equal to $105$. In Fig. \ref{illustration}, the $X$ and the $Y$-axis represent the values taken by the data points and the $Z$-axis represent the value of the function surface at the corresponding data points. From Fig.\ref{illustration}.a it can be seen that for $p=2$ the function $\hat{f}(v)$ is constant everywhere with sharp spikes near the labeled data point. We also illustrate the function surface for $p=2.5$ and $p=\infty$ in Fig.\ref{illustration}.b and Fig. \ref{illustration}.c, respectively.
From Fig. \ref{illustration}.b and Fig. \ref{illustration}.c it can be seen that as the value of $p$ increases the function becomes smooth. This observation led to the formulation of $\ell_{p}$-based Laplacian regularizer problem \cite{lplaplace}: 
\begin{equation}\label{lpgraph}
\begin{array}{ll}
\underset{f}{\rm arg\: min} \:\displaystyle\sum_{i,j \in E} w_{ij}\left|f(v_{i})-f(v_{j})\right|^{p}\\
\hspace{3mm} {\rm{subject\:to}}\: f(v_{i})=h_{i} , i \in L
\end{array}
\end{equation}
The above problem can be cast as an $\ell_{p}$ norm linear regression problem (which we will show in sec. \ref{sec:4}). Hence, the benchmarking algorithm used to solve the problem in (\ref{lpgraph}) is the modified IRLS algorithm \cite{firls}. The paper is organized as follows. In the next section, we discuss the methods used to solve the $\ell_{p}$ norm regression problem and also list the contributions made in this manuscript. In sec. \ref{sec:3}, we give an overview of Majorization Minimization (MM) procedure and in sec. \ref{sec:4} we propose a parallel algorithm to solve the problem in (\ref{eq:12}) using the MM principle for any value of $p$. At the end of the same section we discuss the application of the proposed algorithm for the graph based semi-supervised problem. In sec. \ref{sec:5}, we compare the proposed algorithm with the state-of-the art methods via computer simulations and finally conclude the paper in sec. \ref{sec:6}. 
\section{Related Work and Contributions}\label{sec:2}
The conventional IRLS algorithm used to solve the $\ell_{p}$ norm problem was first developed by Karlovitz \cite{irls4} and also independently by several other researchers (\cite{irls1,irls2,irls5}). A brief history of IRLS algorithm can be found in \cite{irls3}. To compute the value of the next iterate $\bx^{k+1}$, the IRLS algorithm solves the following weighted least-squares problem:
\begin{equation} \label{eq:13}
\begin{array}{ll}
\bx^{k+1} = \quad \underset{\bx}{\rm arg\: min} \: (\bA\bx-\by)^{T} \bW^{k}  (\bA\bx-\by)
\end{array}
\end{equation}
where $\bW$ is a diagonal matrix with diagonal elements $\{w_{1},w_{2} \cdots w_{m}\}$ and $\bW^{k}$ is the value taken by $\bW$ at the $k^{th}$ iteration. The IRLS algorithm starts with $\bW^{0}=\bI$ and solves the problem in (\ref{eq:13}) whose solution is the least-squares estimator $\bx^{0} = \left(\bA^{T}\bA\right)^{-1}\left(\bA^{T}\by\right)$. It then updates the weights matrix $\bW^{k} =\textrm{diag}( |\bA\bx^{k}-\by|^{p-2})$ and solves the weighted least-squares problem in (\ref{eq:13}) whose solution is $\bx^{k} = (\bA^{T}\bW^{k}\bA)^{-1}\bA^{T}\bW^{k}\by$ and this is repeated until convergence. The pseudocode of IRLS algorithm is shown in Table 1. 
\begin{center}
\begin{tabular}{ @{}p{9cm} }
\hline
\hline
\bf{Table 1: Pseudocode of IRLS algorithm} \\
\hline
\hline
{\bf{Input}}: Noisy observed data $\by$, data matrix $\bA$ and $p$ \\
{\bf{Initialize}}: Set \emph{k} = 0. Initialize ${\bx^{0}}=\left(\bA^{T}\bA\right)^{-1}\left(\bA^{T}\by\right)$. \\
{\bf{Repeat}}: \\ 
    1) Update $\bW^{k} =\textrm{diag}( |\bA\bx^{k}-\by|^{p-2})$ \\
    2) $\bx^{k+1}= (\bA^{T}\bW^{k}\bA)^{-1}\bA^{T}\bW^{k}\by$ \\
    $k \leftarrow k+1$\\ 
 \textbf{until} $\left|\dfrac{{f_{_{\textrm{LR}}}({{{\bx}^{k}}})}-{f_{_{\textrm{LR}}}({{\bx}^{k-1}})}}{{f_{_{\textrm{LR}}}({{\bx}^{k-1}})}}\right| \leq 10^{-3}$\\
\hline
\hline
     \end{tabular}
\end{center}
The IRLS algorithm has several drawbacks. Firstly, the IRLS algorithm involves computing the inverse of a square matrix of size $n$ at every iteration - making the algorithm computationally expensive for large $n$. Also, at any iteration, if $\by$ becomes equal to $\bA\bx^{k}$, the matrix $(\bA^{T}\bW^{k}\bA)$ becomes ill-conditioned and the algorithm will run into numerical issues. Another drawback of IRLS algorithm is that it fails to converge for $p>3$ (\cite{irls2},\cite{convergence}). To overcome the issue of convergence, the authors in \cite{irls4} calculated the value of the next iterate by the following weighted combination of the previous iterate $\bx^{k-1}$ and the IRLS iterate $\tilde{\bx}^{k}$ with the updated weights:
\begin{equation}
\begin{array}{ll}
\bx^{k+1} = q\tilde{\bx}^{k}+ (1-q) \bx^{k-1}
\end{array}
\end{equation}
 However, this method is very slow because one has to choose an optimal value for $q$ at every iteration. The authors in \cite{firls} proposed a modified IRLS algorithm wherein to avoid ill-conditioned matrix, they systematically added non-zero values to the diagonal elements of the matrix $\bW^{k}$. Also, to improve the stability of the IRLS algorithm, the next iterate $\bx^{k+1}$ is calculated by doing a line search along the line joining the previous iterate $\bx^{k-1}$ and the standard IRLS iterate with the modified weights. Homotopy based approaches have also been proposed to solve the $\ell_{p}$ norm regression problem \cite{modifiedirls1}, wherein first a simpler optimization problem is solved i.e., the $\ell_{p}$ norm problem for $p=2$ is solved and then $p$ is slowly increased until the desired minimum of the original objective function is reached. The authors in \cite{ip} used interior point methods to solve the $\ell_{p}$ norm problem. However, this method converges very slowly for large dimension of input data \cite{firls}. \\
\\
In this paper, we propose a parallel iterative algorithm to solve the $\ell_{p}$ norm regression problem for any $p$ based on the Majorization Minimization (MM) procedure (which will be introduced briefly in the next section) and show through numerical simulations that the proposed algorithm converges to the stationary point of the problem for any random initialization and enjoys faster speed of convergence. The major contributions of this paper are:
\begin{enumerate}
\item{A MM based parallel algorithm - \textbf{P}arallel Ite\textbf{R}ative Alg\textbf{O}rith\textbf{M} for $\ell_{\textbf{P}}$ norm regression via Majoriza\textbf{T}ion Minimization (\textbf{PROMPT}) is proposed to solve the $\ell_{p}$ norm regression problem for $p \in [1, \infty]$. The proposed algorithm can update each element of $\bx$ parallely - which is useful for solving large dimensional problem.}
\item{We prove that the proposed algorithm converges to the stationary point of the problem.}
\item{We show that the proposed algorithm can be applied for the graph based semi-supervised learning problem.}
\item{Numerical simulations are conducted to compare the proposed algorithm with the state-of-the-art algorithms.}
\end{enumerate}
\section{Majorization Minimization}\label{sec:3}
In this section, we give an overview of MM procedure which has been applied to develop \textbf{PROMPT}. To explain the MM framework, we consider the following optimization problem: 
\begin{equation}
\begin{array}{ll}
\underset{\bx \in \chi}{\rm minimize} \: f(\bx)
\end{array}
\end{equation}
where $f(\bx)$ could be either a convex or non-convex function and $\chi$ is the constraint set. Under the MM framework, at every iteration, a surrogate function $g(\bx|\bx^{k})$ which majorizes the function $f(\bx)$ is constructed and minimized i.e.
\begin{equation}  \label{eq:mmc}
\bx^{k+1} \in \underset{\bx \in \chi}{\rm arg\:min} \quad g\left(\bx|\bx^{k}\right)
\end{equation}
where $\bx^{k+1}$ is the value taken by $\bx$ at the $(k+1)^{th}$ iteration. A surrogate function $g(\bx|\bx^{k})$ is said to majorize a function $f(\bx)$ if it satisfies the following properties:
\begin{equation}  \label{eq:mma}
g\left(\bx^{k}|\bx^{k}\right) = f\left(\bx^{k}\right) 
\end{equation}
\begin{equation}\label{eq:mmb}
g\left(\bx|\bx^{k}\right) \geq f\left(\bx\right) 
\end{equation}
Using (\ref{eq:mmc}), (\ref{eq:mma}) and (\ref{eq:mmb}) it can be shown that the sequence of points $\{\bx^{k}\}$ generated by the MM procedure monotonically decrease the objective function: 
\begin{equation}
\begin{array}{ll}
f(\bx^{k+1}) \leq g(\bx^{k+1}|\bx^{k}) \leq g(\bx^{k}|\bx^{k})  = f(\bx^{k})
\end{array}
\end{equation}
where the first inequality and the last equality are obtained by using (\ref{eq:mma}) and (\ref{eq:mmb}). The second inequality is by (\ref{eq:mmc}). The computational complexity and the convergence speed of the algorithm depends on the choice of the surrogate function. For instance, in the case of a multivariate optimization problem, a surrogate function could make the parameters separable and hence each of them could be updated parallely - which is particularly useful for a large scale problem.  An overview of the various surrogate functions used can be found in (\cite{sir}, \cite{tutorial}).
\section{Proposed Algorithm for $\ell_{p}$ norm linear regression problem}\label{sec:4}
In this section, we first propose an iterative algorithm \textbf{PROMPT} based on the MM principle  to solve the $\ell_{p}$ norm linear regression problem for $p \in (1, \infty)$. The proposed algorithm can parallely update each element of $\bx$ which is useful to handle large scale data efficiently. We then discuss the extensions of the proposed algorithm to solve the $\ell_{1}$ and the $\ell_{\infty}$ norm linear regression problems. Next, we show that the proposed algorithm converges to the stationary point of the $\ell_{p}$ norm linear regression problem. Finally, at the end of the section, we discuss the application of the proposed algorithm for the graph based semi-supervised learning problem.
\subsection{\textbf{P}arallel Ite\textbf{R}ative Alg\textbf{O}rith\textbf{M} for $\ell_{\textbf{P}}$ norm regression via Majoriza\textbf{T}ion Minimization (\textbf{PROMPT})}
\vspace{-1mm}
The objective function $f_{_{LR}}(\bx)$ in (\ref{eq:12}) is not separable in each element of $\bx$, which can be observed by first rewriting $f_{_{LR}}(\bx)$ as:
\begin{equation}\label{eq:31}
\begin{array}{ll}
f_{_{LR}}(\bx) = \|\by-\bA\bx\|_{p} = \displaystyle\sum_{i=1}^{m} |y_{i} - \ba_{i}^{T}\bx|^{p}
\end{array}
\end{equation}
where $\ba_{i} \in \mathbf{R}^{n}$ is the $i^{th}$ row of $\bA$ matrix and $y_{i}$ is the $i^{th}$ element of $\by$. Expanding $|y_{i} - \ba_{i}^{T}\bx|^{p}$, one gets terms coupled in the elements of $\bx$ which makes the parallel minimization of $f_{_{LR}}(\bx)$ challenging. In this section, we develop a parallel algorithm using the MM principle in which we form a surrogate function $g(x_{j}|\bx^{k})$ which majorizes $f_{_{LR}}(\bx)$. The surrogate function is separable in each element of $\bx$ and hence each element of $\bx$ can be updated parallely. To develop the surrogate function we make use of the following lemma:
\vspace{-0.3mm}
\begin{lemma} \label{lemma 1}
Given any $\tilde{\bx} = {\tilde{\bx}^{k}}$, the function $\left|{\bc^{T}}\tilde{\bx}\right|^{p}$ ($p\geq 1$), where $\bc\: \rm{and}\: \tilde{\bx} \in \mathbf{R}^{n+1}$, can be upper bounded as:
\begin{equation}\label{eq:21a}
\begin{array}{ll}
\left|\left({\bc^{T}}\tilde{\bx}\right)\right|^{p} \leq \displaystyle\sum_{j=1}^{n+1}\dfrac{1}{n+1}\left|\left(n+1\right) c_{j}\left(\tilde{x}_{j}-\tilde{x}_{j}^{k}\right)+ \bc^{T}{\tilde{\bx}^{k}}\right|^{p}
\end{array}
\end{equation} 
\vspace{-6.5mm}
\end{lemma}
\begin{IEEEproof}
We replicate the proof from \cite{sir} for the sake of clarity. Note that the function $\left|\left(\cdot\right)\right|^{p}$ ($p \geq 1$) is convex and hence by using the Jensen's inequality we get: 
\begin{equation} \label{jensens}
\begin{array}{ll}
\left|\left(\displaystyle\sum_{j=1}^{n+1} \dfrac{s_{j}}{n+1}\right)\right|^{p} \leq \displaystyle\sum_{j=1}^{n+1}\dfrac{|\left(s_{j}\right)|^{p}}{n+1}
\end{array}
\end{equation}
Letting $s_{j} = (n+1) c_{j}\left(\tilde{x}_{j} - \tilde{x}_{j}^{k}\right) +\bc^{T}(\tilde{\bx}^{k})$ and substituting it in (\ref{jensens}), the inequality in (\ref{eq:21a}) is achieved.
\end{IEEEproof} 
Let $\bc_{i} =[c_{i,1} \cdots c_{i,n+1}]^{T} =  [-\ba_{i}\:\: y_{i}]^{T}$ and let $\tilde{\bx} = [\bx, 1]^{T}$. Then the objective function $f_{_{LR}}(\bx)$ in (\ref{eq:31}) can be rewritten as: 
\begin{equation} 
\begin{array}{ll}
f_{_{LR}}(\bx) = \displaystyle\sum_{i=1}^{m} |y_{i} - \ba_{i}^{T}\bx|^{p} =  \displaystyle\sum_{i=1}^{m} |\bc_{i}^{T}\tilde{\bx}|^{p}
\end{array}
\end{equation}
Using lemma \ref{lemma 1}, we majorize the term $|\bc_{i}^{T}\tilde{\bx}|^{p}$ and hence majorize the objective function $f_{_{LR}}(\bx)$. At any given $\bx^{k}$, the majorization function $g(x_{j}|\bx^{k})$ is given by: 
\begin{equation}
\begin{array}{ll}
g(x_{j}|\bx^{k}) = \displaystyle\sum_{i=1}^{m}\displaystyle\sum_{j=1}^{n+1}\dfrac{1}{n+1}\left|\left(n+1\right) c_{ij}\left(\tilde{x}_{j}-\tilde{x}_{j}^{k}\right)+ \bc_{i}^{T}{\tilde{\bx}^{k}}\right|^{p}
\end{array}
\end{equation}
where $c_{ij}$ represents the $j^{th}$ element of the ${i}^{th}$ vector $\bc_{i}$ . The above surrogate function can be rewritten as:
 \begin{equation}\label{nostuck}
\begin{array}{ll}
g(x_{j}|\bx^{k}) = \displaystyle\sum_{i=1}^{m}\displaystyle\sum_{j=1}^{n}\left|c_{ij}\tilde{x}_{j} +d^{k}_{ij}\right|^{p}+\\ \displaystyle\sum_{i=1}^{m}\left|c_{i, n+1}\tilde{x}_{n+1} +d^{k}_{i, n+1}\right|^{p} = \displaystyle\sum_{i=1}^{m}\displaystyle\sum_{j=1}^{n}\left|-a_{ij}{x}_{j} +d^{k}_{ij}\right|^{p}+\\  \hspace{15mm}\displaystyle\sum_{i=1}^{m}\left|y_{i} +d^{k}_{i, n+1}\right|^{p}
\end{array}
\end{equation}
where $d^{k}_{ij} =  -c_{ij}\tilde{x}^{k}_{j}+ \dfrac{1}{n+1}\bc_{i}^{T}\bx^{k}$. From above, it can be observed that if $a_{ij}$ is zero then the product $a_{ij}x_{j}$ becomes equal to zero, hence the first term of $g(x_{j}|\bx^{k})$ has to be computed only over the non-zero elements of $a_{ij}$ i.e., 
\begin{equation}\label{surrogate_1}
\begin{array}{ll}
g(x_{j}|\bx^{k}) = \displaystyle\mathop{\sum_{i=1}^{m}\sum_{j=1}^{n}}_{a_{ij}\neq 0}\left|-a_{ij}{x}_{j} +d^{k}_{ij}\right|^{p} + \displaystyle\sum_{i=1}^{m}\left|y_{i} +d^{k}_{i, n+1}\right|^{p}
\end{array}
\end{equation}
Ignoring the constant terms in $g(x_{j}|\bx^{k})$, the surrogate minimization problem at any iteration, given $\bx^{k}$ becomes: 
 \begin{equation}\label{surrogate1}
\begin{array}{ll}
\underset{\bx}{\rm arg\: min} \displaystyle\mathop{\sum_{i=1}^{m}\sum_{j=1}^{n}}_{a_{ij}\neq 0}w_{ij}\left|{x}_{j} - \dfrac{d^{k}_{ij}}{a_{ij}} \right|^{p} 
\end{array}
\end{equation}
where $w_{ij} = |a_{ij}|^{p}$. The above problem is separable in each element of $\bx$. Hence, at every iteration we solve the following surrogate minimization problem parallely over each element of $\bx$: 
 \begin{equation}\label{surrogate}
\begin{array}{ll}
\underset{x_{j}}{\rm arg\: min} \displaystyle\mathop{\sum_{i=1}^{m}}_{a_{ij}\neq 0}w_{ij}\left|{x}_{j} - \dfrac{d^{k}_{ij}}{a_{ij}} \right|^{p} 
\end{array}
\end{equation}
Therefore, each element of $\bx$ can be updated parallely. The problem in (\ref{surrogate}) does not have a closed-form solution. However, we can employ a parameter free bisection method to solve the above problem. The initial interval $[a,b]$ of the bisection method for the problem in (\ref{surrogate}) is found using the following lemma:
\vspace{-1mm}
\begin{lemma} \label{lemma 2}
Consider the following problem wherein given the positive weights $(w_{1}, w_{2} \cdots w_{N})$ and the data points $(a_{1}, a_{2} \cdots a_{N})$, the problem is to estimate $x$ such that it minimizes the weighted sum of absolute deviations raised to the $p^{th}$ power, i.e. 
\begin{equation}\label{simplified}
\begin{array}{ll}
\underset{x}{\rm arg\: min} \: f(x) = \displaystyle\sum_{j=1}^{N} w_{j} |x-a_{j}|^{p}
\end{array}
\end{equation}
The solution $x$ of the above problem will always lie in-between $\rm{min}(\ba)$ and $\rm{max}(\ba)$ where $\ba = [a_{1},a_{2} \cdots a_{N}]^{T}$. 
\end{lemma}
\begin{IEEEproof}
The gradient of the problem in (\ref{simplified}) is given as:
\begin{equation}
\begin{array}{ll}
f'(x) = \displaystyle\sum_{j=1}^{N} p\:w_{j} |x-a_{j}|^{p-2} (x-a_{j}) = \displaystyle\sum_{j=1}^{N}v_{j} (x-a_{j}) 
\end{array}
\end{equation}
where $v_{j} =p\:w_{j} |x-a_{j}|^{p-2}$  and is always positive. Note that for $x= \rm{min}(\ba)$ and $x = \rm{max}(\ba)$, the gradient $f'(x)$ will always be lesser than and greater than zero, respectively. Then, according to the Intermediate Value Theorem \cite{ivt}, the gradient will be equal to zero for $x$ in the interval  [$\rm{min}(\ba)$, $\rm{max}(\ba)$].
\end{IEEEproof}
Using lemma \ref{lemma 2}, the initial interval $[a,b]$ for the bisection search approach to solve the problem in (\ref{surrogate}) is $[\rm{min}(\bs_{j})\:\:\rm{max}(\bs_{j})]^{T}$ where the $i^{th}$ element of $\bs_{j}$ is given as $s_{ij} = \dfrac{d^{k}_{ij}}{a_{ij}} $. Further, by the principle of MM, it is sufficient that the bisection search approach finds a value $\bx^{k}$ at the $k^{th}$ iteration such that $f_{_{\textrm{LR}}}({\bx}^{k}) <f_{_{\textrm{LR}}}({\bx}^{k-1})$ and not find the exact minimum of the surrogate minimization problem in (\ref{surrogate}).
Pseudocode of the proposed algorithm is shown in Table 2. We would like to point out here that unlike the IRLS algorithm, the proposed algorithm does not have the problem of getting stuck - since the minimum in (\ref{surrogate}) is found over the non-zero elements of $a_{ij}$'s. Also, when compared to IRLS and the modified IRLS algorithm, the proposed algorithm does not involve computing matrix inverse. 
\begin{center}
\begin{tabular}{ p{8cm} }
\hline
\hline
\bf{Table 2: Pseudocode of the proposed algorithm for $\ell_{p}$ norm linear regression problem for $p \in (1, \infty)$.} \\
\hline
\hline
{\bf{Input}}: Noisy observed data $\by$, data matrix $\bA$ and $p$  \\
{\bf{Initialize}}: Set $k=0$. Initialize ${{\bx}^{0}}$  \\    
{\bf{Precompute}}:  $w_{ij} = |a_{ij}|^{p}$, $i \in (1, \cdots m)$ and $j \in (1, \cdots n)$\\         
{\bf{Repeat}}: \\
Compute the following parallely over all the elements of $\bx$: \\
1)\, Compute $s_{ij} = \dfrac{d^{k}_{ij}}{a_{ij}} = x^{k}_{j} + \dfrac{y_{i} - \ba_{i}^{T}\bx^{k}}{(n+1)a_{ij} }$,  $i \in (1, 2, \cdots m)$.\\
2)\, $x_{j}^{k+1}$ is obtained by solving (\ref{surrogate}) using bisection method with $a = \rm{min}\left(\bs_{j}\right)$ and $b =\rm{max}\left(\bs_{j}\right)$.\\
 $k \leftarrow k+1$, \textbf{until} $\left|\dfrac{{f_{_{\textrm{LR}}}({{{\bx}^{k}}})}-{f_{_{\textrm{LR}}}({{\bx}^{k-1}})}}{{f_{_{\textrm{LR}}}({{\bx}^{k-1}})}}\right| \leq 10^{-3}$\\
\hline
\hline
\end{tabular}
\end{center}
Before ending this subsection, we will here discuss the  computational complexity of the proposed algorithm. First, we would like to point out that the weights $w_{ij}$ which are required to compute $x_{j}^{k+1}$ are independent of the optimization variable and hence can be precomputed with a complexity of $\mathcal{O}(p)$. Next, at every iteration, the complexity of the algorithm is only dictated by the computation of the inner product $\ba_{i}^{T}\bx^{k}$ and the bisection search step. The complexity of the inner product $\ba_{i}^{T}\bx^{k}$ is $\mathcal{O}(n)$ and the  complexity of the bisection search algorithm is $\mathcal{O}({\textrm{log}}\:l)$, where $l$ is the number of sub-intervals. Hence, the total complexity of the proposed algorithm is $\mathcal{O}\left(p+ k\left(n+{\textrm{log}}\:l\right)\right)$. Also as mentioned earlier, the complexity of the state-of-the-art IRLS algorithm is dictated by the computation of the inverse of the  matrix $\bA^{T}\bW^{k}\bA$ and hence the complexity of the IRLS algorithm is $\mathcal{O}(n^{3})$. Hence, the proposed algorithm has the least computational complexity when compared to the state-of-the-art algorithm.
\subsection{Special Cases}
When compared to the $\ell_{p}$ norm regression problem for $p \in (1, \infty)$, the $\ell_{1}$ and $\ell_{\infty}$ norm regression problems are non-differentiable - thereby one cannot use the bisection search algorithm developed in the previous subsection to solve the $\ell_{1}$ and $\ell_{\infty}$ norm regression problem. Hence, in this subsection, we discuss an alternate approach to solve the surrogate minimization problem and extend PROMPT algorithm to solve the $\ell_{1}$ and $\ell_{\infty}$ norm regression problems.\\
1. {$\ell_{1}$ norm linear regression} - Consider the following $\ell_{1}$ norm linear regression problem: 
\begin{equation}\label{eq:l1}
\begin{array}{ll}
f_{_{LR}}(\bx) = \|\by-\bA\bx\|_{1} = \displaystyle\sum_{i=1}^{m} |y_{i} - \ba_{i}^{T}\bx|
\end{array}
\end{equation}  
Similar to the $\ell_{p}$ norm regression problem ($p>2$) in (\ref{eq:31}), we rewrite the above problem as:
\begin{equation}\label{eq:l1-simp}
    \begin{array}{ll}
    f_{_{LR}}(\bx)=\displaystyle\sum_{i=1}^{m} |\bc_{i}^{T}\tilde{\bx}|
    \end{array}
\end{equation}
where $\bc_{i} =[c_{i,1} \cdots c_{i,n+1}]^{T} =  [-\ba_{i}\:\: y_{i}]^{T}$ and let $\tilde{\bx} = [\bx, 1]^{T}$. Then by using lemma \ref{lemma 1}, we majorize the term $|\bc_{i}^{T}\tilde{\bx}|$ and obtain the following surrogate minimization problem: 
 \begin{equation}\label{surrogate-l1}
\begin{array}{ll}
\underset{x_{j}}{\rm arg\: min} \displaystyle\mathop{\sum_{i=1}^{m}\sum_{j=1}^{n}}w_{ij}\left|{x}_{j} - s_{ij} \right| 
\end{array}
\end{equation}
where $w_{ij}=|a_{ij}|$, $s_{ij}=\dfrac{d^{k}_{ij}}{a_{ij}}$ and $d^{k}_{ij} =  -c_{ij}\tilde{x}^{k}_{j}+ \dfrac{1}{n+1}\bc_{i}^{T}\bx^{k}$. Similar to the surrogate mimization problem in (\ref{surrogate}), the above problem is separable in each element of $\bx$ and hence can be updated parallely. In addition, unlike the surrogate minimization problem in (\ref{surrogate}), the above surrogate minimization problem has a closed-form solution and is given by the weighted median of the points $s_{ij}$ \cite{Wmed}. Given the points $s_{1j}, s_{2j},\cdots, s_{mj}$ and its associated weights $w_{ij}$, the weighted median $x_{j}$ is calculated as\\
a. Sort the points $s_{ij}$ in ascending order.\\
b. Normalize the weights $w_{ij}$ associated with the data points $s_{ij}$ such that the sum of the weights is equal to one.\\
c. The weighted median $x_{j}$ is one of the points $s_{qj}$ satisfying $\displaystyle\sum_{i=1}^{q-1}w_{ij} \leq \dfrac{1}{2}$ and $\displaystyle\sum_{i=q+1}^{m}w_{ij} \leq \dfrac{1}{2}$.\\
The pseudocode of the proposed algorithm for $\ell_{1}$ norm regression problem is shown in Table 3. 
\vspace{5mm}
\begin{center}
\begin{tabular}{ p{8cm} }
\hline
\hline
\bf{Table 3: Pseudocode of the proposed algorithm for $\ell_{1}$ norm linear regression problem} \\
\hline
\hline
{\bf{Input}}: Noisy observed data $\by$, data matrix $\bA$ \\
{\bf{Initialize}}: Set $k=0$. Initialize ${{\bx}^{0}}$  \\    
{\bf{Precompute}}:  $w_{ij} = |a_{ij}|$, $i \in (1, \cdots m)$ and $j \in (1, \cdots n)$\\         
{\bf{Repeat}}: \\
Compute the following parallely over all the elements of $\bx$: \\
1)\, Compute $s_{ij} = \dfrac{d^{k}_{ij}}{a_{ij}} = x^{k}_{j} + \dfrac{y_{i} - \ba_{i}^{T}\bx^{k}}{(n+1)a_{ij} }$,  $i \in (1, 2, \cdots m)$.\\
\hline
\hline
\end{tabular}
\end{center}
\begin{center}
\begin{tabular}{ p{8cm} }
\hline
\hline
\bf{Table 3: Pseudocode of the proposed algorithm for $\ell_{1}$ norm linear regression problem} \\
\hline
\hline
2)\, $x_{j}^{k+1}$ is obtained by finding the weighted median of the points $s_{ij}$ as discussed in Subsection \ref{sec:4}.B. \\
 $k \leftarrow k+1$, \textbf{until} $\left|\dfrac{{f_{_{\textrm{LR}}}({{{\bx}^{k}}})}-{f_{_{\textrm{LR}}}({{\bx}^{k-1}})}}{{f_{_{\textrm{LR}}}({{\bx}^{k-1}})}}\right| \leq 10^{-3}$\\
 \hline
\hline
\end{tabular}
\end{center}
2. {$\ell_{\infty}$ norm linear regression} - The $\ell_{\infty}$ norm linear regression problem is given as: 
\begin{equation}\label{eq:linfty}
\begin{array}{ll}
\underset{\bx}{\rm arg\: min} \{f_{_{LR}}(\bx) = \|\by-\bA\bx\|_{\infty} =\\ \hspace{25mm}\displaystyle\max_{1,2,\cdots,m} |y_{i} - \ba_{i}^{T}\bx|\}
\end{array}
\end{equation}
Similar to the development of the PROMPT algorithm, we majorize the objective function in (\ref{eq:linfty}) using lemma \ref{lemma 1} and obtain the following surrogate minimization problem:
\begin{equation}\label{eq:linfty-surrogate}
\begin{array}{ll}
\underset{x_{j}}{\rm min}   \displaystyle\max_{1,2,\cdots,m}w_{ij}\left|{x}_{j} - s_{ij} \right| 
\end{array}
\end{equation}
Note that the above surrogate minimization problem is separable in each element of $\bx$ and therefore we can solve the above problem parallely over each element of $\bx$:
To solve the above problem, we first rewrite the problem in the epigraph form as: 
\begin{equation}\label{eq:epi}
\begin{array}{ll}
\underset{x_{j},z}{\rm min}\:\:  z\\
\textrm{subject to}\:\: w_{ij}^{2} \left|{x}_{j} - s_{ij} \right|^{2}  \leq z\:\: i=1,2,\cdots,m
\end{array}
\end{equation}
The Lagrange of the above problem is:
\begin{equation}
    \begin{array}{ll}
L(x_{j},z,\lambda) = z + \displaystyle\sum_{i=1}^{m} \lambda_{i}\left(w_{ij}^{2} \left|{x}_{j} - s_{ij} \right|^{2}-z\right)
    \end{array}
\end{equation}
where $\lambda_{i}$'s are the Lagrange multiplier associated with each inequality. Next, we derive the KKT conditions for the problem in (\ref{eq:epi}). 
\vspace{-1.5mm}
\begin{enumerate}
\begin{item}Minimizing the Lagrange function with respect to $z$ we get: 
\begin{equation}\label{c1}
    \begin{array}{ll}
    \displaystyle\sum_{i=1}^{m}\lambda_{i}=1
    \end{array}
\end{equation}
\end{item}
\begin{item}
Minimizing the Lagrange function with respect to $x_{j}$ we get: 
\begin{equation}
    \begin{array}{ll}
    x_{j} = \dfrac{\displaystyle\sum_{i=1}^{m} \lambda_{i} w_{ij}^{2}s_{ij}}{\displaystyle\sum_{i=1}^{m}\lambda_{i}w_{ij}^{2}}
    \end{array}
\end{equation}
\end{item}
\begin{item}
By the complementary slackness property we have: 
\begin{equation}
    \begin{array}{ll}
    \lambda_{i}\left(z-w_{ij}^{2}|x_{j}-s_{ij}|^{2}\right) =0
    \end{array}
\end{equation}
\end{item}
\begin{item} Each dual variable $\lambda_{i}$ must be non-negative. 
\end{item}
\vspace{-1.5mm}
Since the problem in (\ref{eq:epi}) is convex, any pair of $(x_{j}, \lambda_{i})$ that satisfies the KKT conditions would be the primal and dual optimal solutions. Hence, we use the above KKT conditions to  solve the problem in (\ref{eq:epi}) iteratively \cite{linf}. The pseudo code of the proposed algorithm to solve the $\ell_{\infty}$ norm linear regression problem is shown in Table. 3. 
\begin{center}
\begin{tabular}{ p{7.9cm} }
\hline
\hline
\bf{Table 3: Pseudocode of the proposed algorithm for $\ell_{\infty}$ norm linear regression problem} \\
\hline
\hline
{\bf{Input}}: Noisy observed data $\by$, data matrix $\bA$ \\
{\bf{Initialize}}: Set $k=0$. Initialize ${{\bx}^{0}}$ and ${{\lambda}_{i}^{0}}$.  \\    
{\bf{Precompute}}:  $w_{ij} = |a_{ij}|$, $i \in (1, \cdots m)$ and $j \in (1, \cdots n)$\\         
{\bf{Repeat}}: \\
Compute the following parallely over all the elements of $\bx$: \\
1)\, Compute $s_{ij} = \dfrac{d^{k}_{ij}}{a_{ij}} = x^{k}_{j} + \dfrac{y_{i} - \ba_{i}^{T}\bx^{k}}{(n+1)a_{ij} }$,  $i \in (1, 2, \cdots m)$.\\
2)\, Using the KKT conditions solve the problem in (\ref{eq:epi}) iteratively to obtain $x_{j}^{k+1}$ \\
 $k \leftarrow k+1$, \textbf{until} $\left|\dfrac{{f_{_{\textrm{LR}}}({{{\bx}^{k}}})}-{f_{_{\textrm{LR}}}({{\bx}^{k-1}})}}{{f_{_{\textrm{LR}}}({{\bx}^{k-1}})}}\right| \leq 10^{-3}$\\
\hline
\hline
\end{tabular}
\end{center}
\end{enumerate}
\subsection{Proof of convergence}
Given that the proposed algorithm is based on MM procedure, it is guaranteed that the sequence of points $\{{\bx}^{k}\}$ generated by MM algorithm will monotonically decrease the problem in (\ref{eq:12}). Moreover, since $f_{_{\textrm{LR}}}({\bx})$ in (\ref{eq:12}) is bounded below by zero, it is ensured that the sequence $f_{_{\textrm{LR}}}({{\bx}})$ will converge to a finite value. \\
We now show that the sequence $\{{\bx}^{k}\}$ converges to the stationary point of the problem in (\ref{eq:12}). From the monotonic property of MM we have: 
\begin{equation}\label{eq:conv}
\begin{array}{ll}
f_{_{\textrm{LR}}}({{\bx}^{0}})\geq f_{_{\textrm{LR}}}({{\bx}^{1}})\geq f_{_{\textrm{LR}}}({{\bx}^{2}})
\end{array}
\end{equation}
Assume that there is a subsequence ${\bx}^{r_{j}}$ converging to a limit point $\tilde{\bq}$. Then from (\ref{eq:mma}), (\ref{eq:mmb}) and  (\ref{eq:conv}) we get:
\begin{equation}
\begin{array}{ll}
g({\bx}^{r_{j+1}}|{\bx}^{r_{j+1}}) = f_{_{\textrm{LR}}}({{\bx}^{r_{j+1}}}) \leq f_{_{\textrm{LR}}}({{\bx}^{r_{j}+1}}) \leq g({\bx}^{r_{j}+1}|{\bx}^{r_{j}}) \\\hspace{25mm}\leq g({\bx}|{\bx}^{r_{j}})
\end{array}
\end{equation}
where $g(\cdot)$ is the surrogate function as defined in (\ref{surrogate_1}). Then, letting $j \rightarrow \infty$, we get:
\begin{equation}
\begin{array}{ll}
g(\tilde{\bq}|\tilde{\bq})  \leq g({\bx}|\tilde{\bq})
\end{array}
\end{equation}
which implies $g'(\tilde{\bq}|\tilde{\bq}) \geq 0$. Since the first order behavior of surrogate function is same as function $f_{_{\textrm{LR}}}({{\bx}})$ (\cite{convergence}), $g'(\tilde{\bq}|\tilde{\bq}) \geq 0$  implies $f'_{_{\textrm{LR}}}({\tilde{\bq}}) \geq 0$. Hence, $\tilde{\bq}$ is the stationary point of $f_{_{\textrm{LR}}}({{\bx}})$ and therefore the proposed algorithm converges to the stationary point of the problem in (\ref{eq:12}).
\subsection{Graph Based semi-supervised learning}
In this subsection we discuss the application of proposed algorithm for the graph based semi-supervised learning problem. Graph based semi-supervised learning consists of two steps:
\begin{enumerate}
\item{Construction of graph $G=(V,E)$ from the given data.} 
\item{Using the constructed graph estimating the labels of the unlabeled data points using an appropriate algorithm.} 
\end{enumerate}
We now discuss each step in detail. In the case of first step, as suggested in \cite{convergence}, one can construct a symmetric $K$-nearest neighbour (K-NN) graph from the given data. In K-NN graph, each vertex are used to represent a data point. An edge between the data points $\bv_{i}$ and $\bv_{j}$ exists if and only if $\bv_{i} \in {\textrm{KNN}}(\bv_{j})$ or $\bv_{j} \in {\textrm{KNN}}(\bv_{i})$, where ${\textrm{KNN}}(\bv_{j})$ stands for the $K$-nearest neighbours of the $\bv_{j}$ among the data points $\bv_{1},\cdots,\bv_{j-1},\bv_{j+1},\cdots,\bv_{n}$. The $K$-nearest neighbours of $\bv_{j}$ are found by measuring the Euclidean distance between $\bv_{j}$ and each of the remaining data points $\bv_{1},\cdots,\bv_{j-1},\bv_{j+1},\cdots,\bv_{n}$ and choosing the data points corresponding to the $K$ smallest distance. Each edge is associated with a non-negative edge weight which is defined as:
\begin{eqnarray}\label{weights}
w_{ij} & = & \begin{cases}
\textrm{exp}\left(-\dfrac{\|\bv_{i}-\bv_{j}\|_{2}^{2}}{\sigma^{2}}\right) & ; \bv_{i} \in {\textrm{KNN}}(\bv_{j})  \\
\textrm{exp}\left(-\dfrac{\|\bv_{j}-\bv_{i}\|_{2}^{2}}{\sigma^{2}}\right) &; \bv_{j} \in {\textrm{KNN}}(\bv_{i}) \\
0 & ;\textrm{otherwise}
\end{cases}\label{eq:toep}
\end{eqnarray}
where $\sigma = \dfrac{1}{2}{\rm{max}}\{\|\bv_{i}-\bv_{j}\|_{2}^{2}: w_{ij}>0\}$. \\
From the constructed graph $G$, we now estimate the labels of the unlabeled points by solving the problem in (\ref{lpgraph}). For the ease of readability, we re-write the problem once again here:
\begin{equation}\label{re}
\begin{array}{ll}
\underset{f}{\rm arg\: min} \:\displaystyle\sum_{i,j \in E} w_{ij} \left|f(v_{i})-f(v_{j})\right|^{p}\\
\hspace{3mm} {\rm{subject\:to}}\: f(v_{i}) = h_{i} , i \in L
\end{array}
\end{equation}
Let $f_{i} = f(v_{i})$ and $\boldf = [f_{1}, f_{2} \cdots f_{n}] \in \mathbf{R}^{n}$ and $\bh = [h_{1}, h_{2}, \cdots, h_{l}] \in \mathbf{R}^{l}$. Then the constrained problem in (\ref{re}) can be re-written as: 
\begin{equation}\label{uc}
\begin{array}{ll}
\underset{\boldf}{\rm arg\: min} \:\displaystyle\sum_{i=1}^{u} \displaystyle\sum_{j=1}^{u} w_{ij} \left|f_{i}-f_{j}\right|^{p} + \displaystyle\sum_{i=1}^{u}  \displaystyle\sum_{j=1}^{l}w_{i,j+u} |f_{i} -h_{j}|^{p}\end{array}
\end{equation}
We now convert the above problem to $\ell_{p}$ norm regression problem. To do so, we first define the edge vertex incidence matrix $\bB$. Each row of the edge incidence matrix $\bB$ corresponds to an edge and each column corresponds to a vertex and its entries are given as:
\begin{eqnarray}
\bB_{ev} & = & \begin{cases}
1 & ;  {\textrm{if v is the head of e}}  \\
-1 &;  {\textrm{if v is the tail of e}}\\
0 & ;\textrm{otherwise}
\end{cases}
\end{eqnarray}
Let $\bG= \bW^{1/p}\bB$, where $\bW$ is a diagonal matrix with the weights of the edges and $\bz = -[\bB]_{:,u+1:u+l}\bh$, where the notation $[\bB]_{:,u+1:u+l}$ is used to represent all the rows of $u+1$ to $u+l$ columns of the matrix $\bB$. Then the problem in (\ref{uc}) can be re-written as:
\begin{equation}
    \begin{array}{ll}
\underset{\boldf}{\rm arg\: min} \|\bG\boldf-\bz\|_{p}
    \end{array}
\end{equation}
which is an $\ell_{p}$-norm regression problem and the proposed algorithm can be used to solve it. 
\section{Numerical simulation}\label{sec:5}
\begin{figure}[!h]
\centering
\begin{subfigure}{0.49\textwidth}
\centering
\captionsetup{justification=centering}
\includegraphics[height=2.4in,width=3.3in]{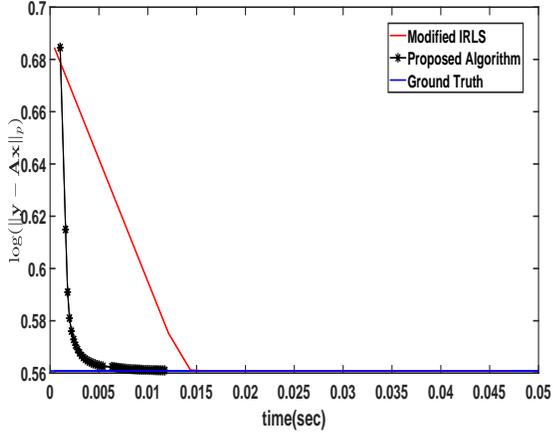} 
\caption{Objective value vs run time of the proposed algorithm and the modified IRLS algorithm}
\end{subfigure}
\begin{subfigure}{0.49\textwidth}
\centering
\captionsetup{justification=centering}
\includegraphics[height=2.4in,width=3.3in]{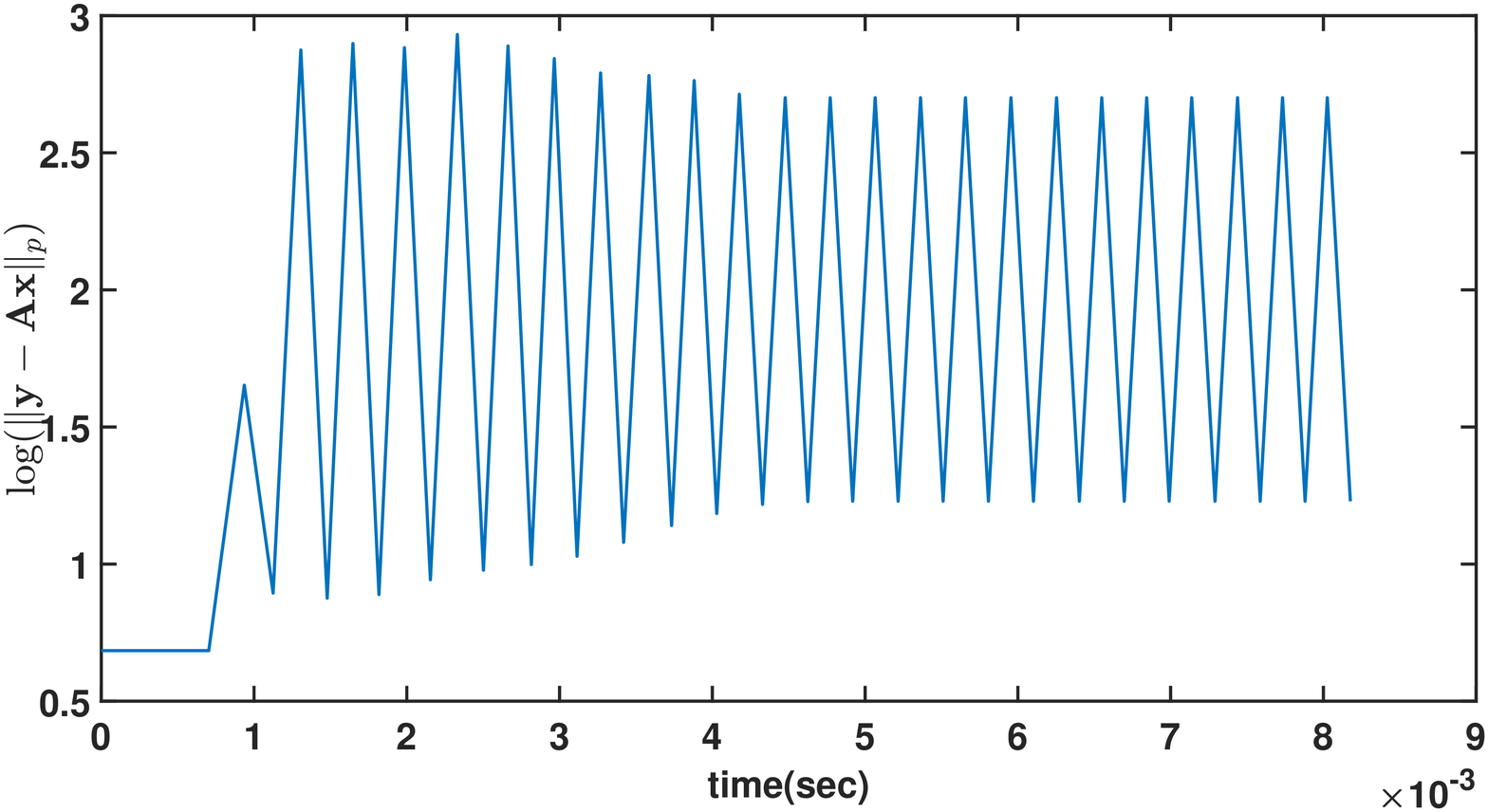}
\caption{Objective value vs run time of the IRLS algorithm}
\end{subfigure}
\caption{Comparison of convergence of the proposed algorithm with the modified IRLS and the IRLS algorithm with $\bx^{0}$ equal to the optimal point of the $\ell_{2}$ norm regression problem.}
\label{fig1}
\end{figure}

\begin{figure}[h]
\centering
\begin{subfigure}{0.49\textwidth}
\centering
\captionsetup{justification=centering}
\includegraphics[height=2.1in,width=3.3in]{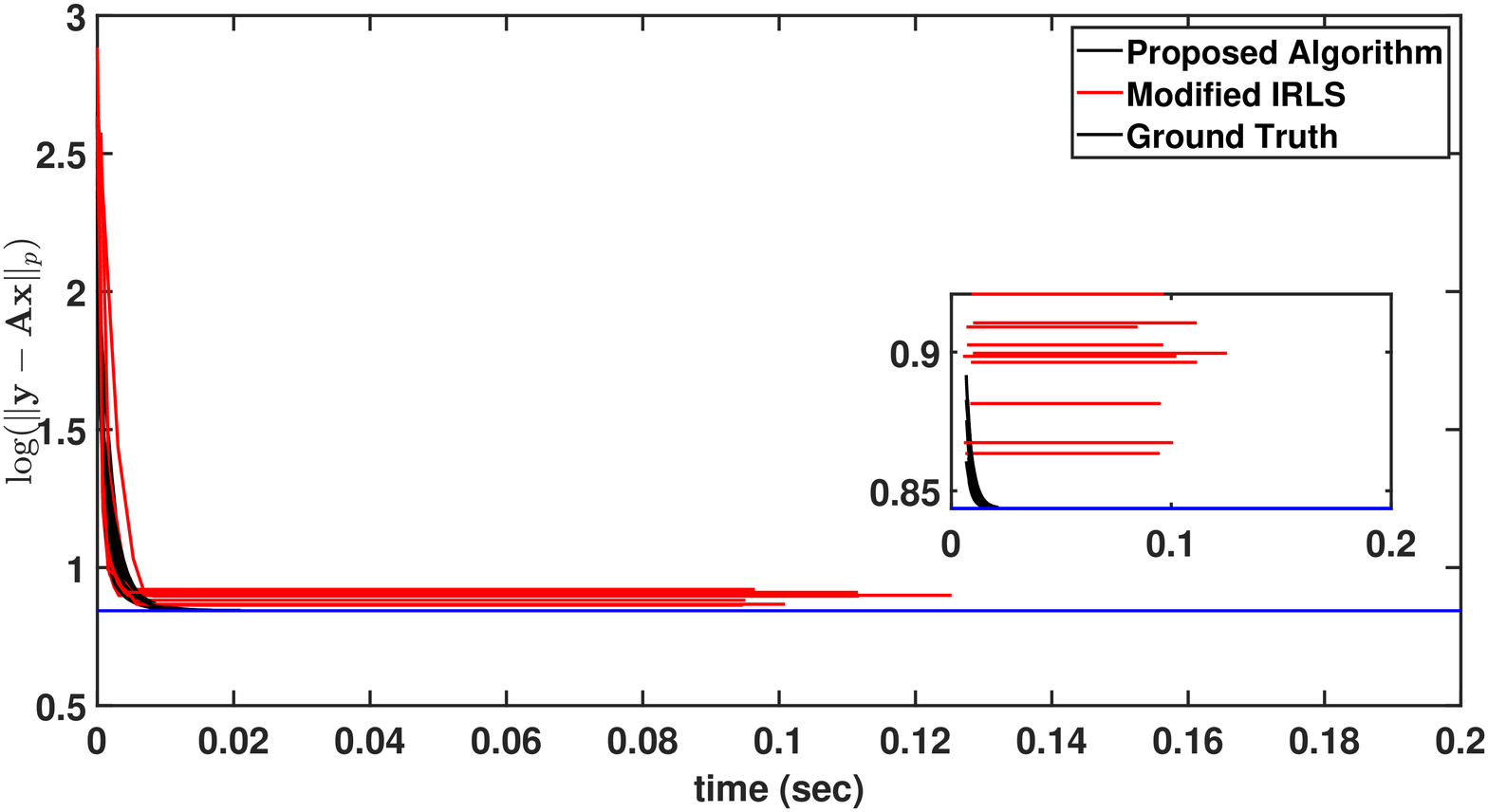}
\caption{Objective value vs run time of the proposed and the algorithm in \cite{firls}}
\end{subfigure}
\begin{subfigure}{0.49\textwidth}
\centering
\captionsetup{justification=centering}
\includegraphics[height=2.1in,width=3.3in]{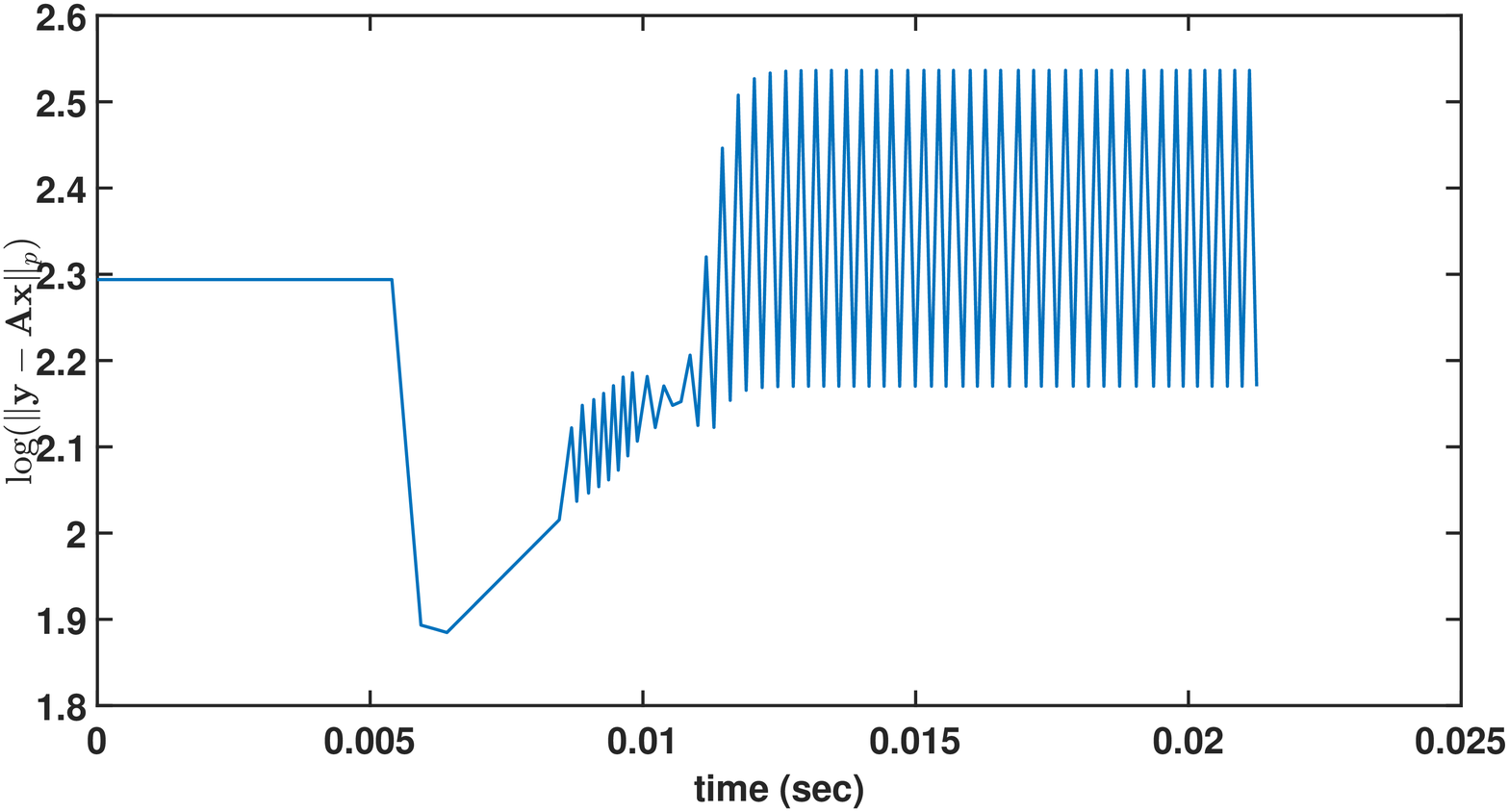}
\caption{Objective value vs run time of the IRLS algorithm}
\end{subfigure}
\caption{Comparison of convergence of the proposed algorithm with the modified IRLS and the IRLS algorithm with $\bx^{0}$ generated  randomly from Normal distribution}
\label{random}
\end{figure}

In this section, we compare the performance of the proposed algorithm with the state-of-the art algorithms used to solve the $\ell_{p}$ norm linear regression problem. In particular, we compare the proposed algorithm with the IRLS algorithm (\cite{irls4, irls1,irls2, irls5}) and the modified IRLS algorithm \cite{firls}. Also, we compare the performance of the proposed algorithm with the modified IRLS algorithm for the Graph based semi-supervised learning problem. The algorithms were implemented in MATLAB. All the simulations were carried out on a PC with $2.40$ GHz Intel Xeon Processor with $64$ GB RAM. \\
\\
1. In this simulation, we fix $p$ equal to $5$, the dimension of the observation vector $\by \in \mathbb{R}^{m}$ as $50$, the dimension of $\bx \in \mathbb{R}^{n}$ as $20$ and compare the
convergence of the proposed algorithm with the state-of-the art algorithms, IRLS algorithm (\cite{irls4, irls1,irls2, irls5}) and the modified IRLS algorithm \cite{firls}. The elements of the matrix $\bA$ and the observed data $\by$ were randomly generated from Normal distribution with zero mean and unit variance. All the algorithms were initialized at ${\bx^{0}}=\left(\bA^{T}\bA\right)^{-1}\left(\bA^{T}\by\right)$, i.e., at the optimal solution of $\ell_{2}$ norm regression problem. Fig. \ref{fig1}.a. shows the objective value vs run time of the proposed algorithm and the modified IRLS algorithm and Fig. \ref{fig1}.b. shows the objective value vs run time of the IRLS algorithm. The ground truth shown in Fig. \ref{fig1}.a. is obtained using CVX \cite{cvx}. From Fig. \ref{fig1} it can be seen that the proposed algorithm and the modified IRLS algorithm converges to the true solution, while the IRLS algorithm fails to converge. Also, from Fig. \ref{fig1}.a. it can be seen that the proposed algorithm converges faster than the  modified IRLS algorithm which could be because the proposed algorithm can parallely update each elements of $\bx$ when compared to the non-parallel modified IRLS algorithm. Although we show the convergence of the algorithms for single run in Fig. \ref{fig1}, we observed similar behavior for many runs.  \\
\\
2. In this simulation, we compare the convergence of the algorithms with the elements of $\bx^{0}$ generated randomly from Normal distribution with mean zero and unit variance. Similar to the previous experiment, we fix $p$ equal to $5$, the dimension of the observation vector $\by$ as $50$ and the dimension of $\bx$ equal to $20$ and generate the elements of the data matrix $\bA$ and the observed data $\by$ randomly from Normal distribution with zero mean and unit variance. Fig. \ref{random}.a. shows the objective value vs run time of the proposed algorithm and the modified IRLS algorithm for $10$ different random initializations $\bx^{0}$. Fig. \ref{random}.b. shows the objective value vs run time of the IRLS algorithm for one such random initialization. From Fig. \ref{random} it can be seen that the proposed algorithm converges to the ground truth while the IRLS and the modified IRLS algorithms fails to converge. Hence, the modified IRLS algorithm converges to the optimal solution of the $\ell_{p}$ norm regression problem only when initialized with the solution of $\ell_{2}$ norm regression problem. This initialization scheme has two issues. Firstly, it requires the matrix $\bA^{T}\bA$ to be invertible and secondly, the computational complexity associated with the inversion of $\bA^{T}\bA$ is $\mathcal{O}(n^{3})$ - which make this initialization expensive for large value of $n$. \\
\\
3. In this simulation, we show that the proposed algorithm converges to the optimal solution irrespective of the initialization. To show the same we generate elements of $\bx^{0}$ from three different distributions -  normal distribution with zero mean and unit variance, uniform distribution from $[0, 1]^{n}$ and exponential distribution with mean equal to $0.1$. Similar to the previous experiments, the dimension of the observation vector and the dimension of $\bx$ were set equal to $50$ and $20$, respectively and their elements were randomly generated from Normal distribution with zero mean and unit variance. The algorithm was made to run until the following condition was met: 
\begin{equation}\label{condition}
\begin{array}{ll}
f_{_{LR}}(\bx^{k}) - f_{_{LR}}(\bx^{*}) \leq 10^{-3} 
\end{array}
\end{equation}
where $f_{_{LR}}(\bx^{*})$ is the optimal value of the problem in (\ref{eq:12}) and is obtained by solving the problem using CVX. Fig. 4 shows the objective value vs run time of the proposed algorithm with the elements of $\bx^{0}$ generated randomly from different distributions and for different values of $p$.  From Fig. 4 it can be seen that the proposed algorithm converges to the optimal solution irrespective of the initialization scheme. Hence, unlike the modified IRLS algorithm, the proposed algorithm is not sensitive to the initialization. 
\begin{figure}[!h]
\centering
\begin{subfigure}{0.49\textwidth}
\centering
\captionsetup{justification=centering}
\includegraphics[height=2.1in,width=3.3in]{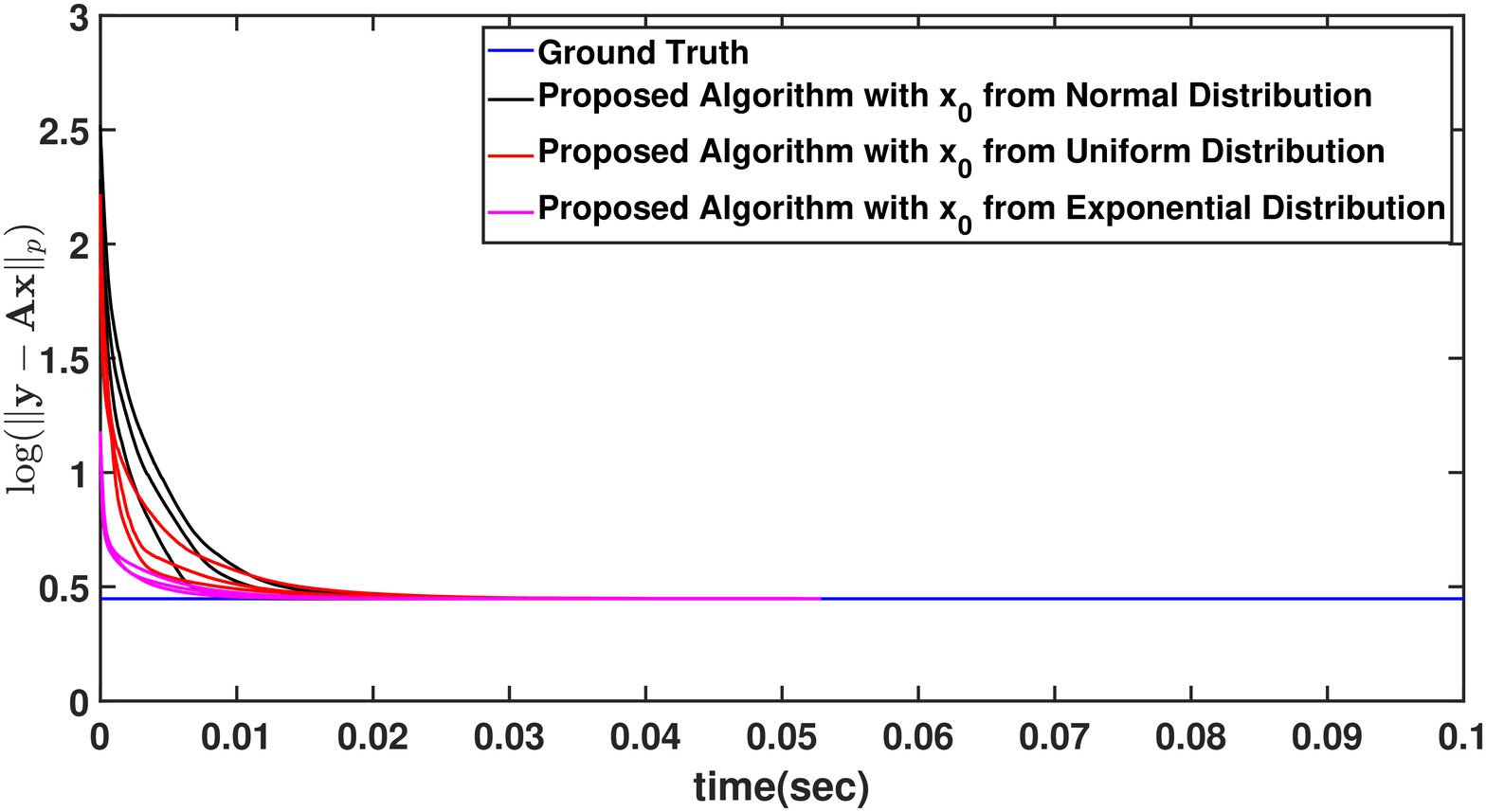}
\caption{Objective value vs run time of the proposed algorithm for $p=10$}
\end{subfigure}
\begin{subfigure}{0.49\textwidth}
\centering
\captionsetup{justification=centering}
\includegraphics[height=2.1in,width=3.3in]{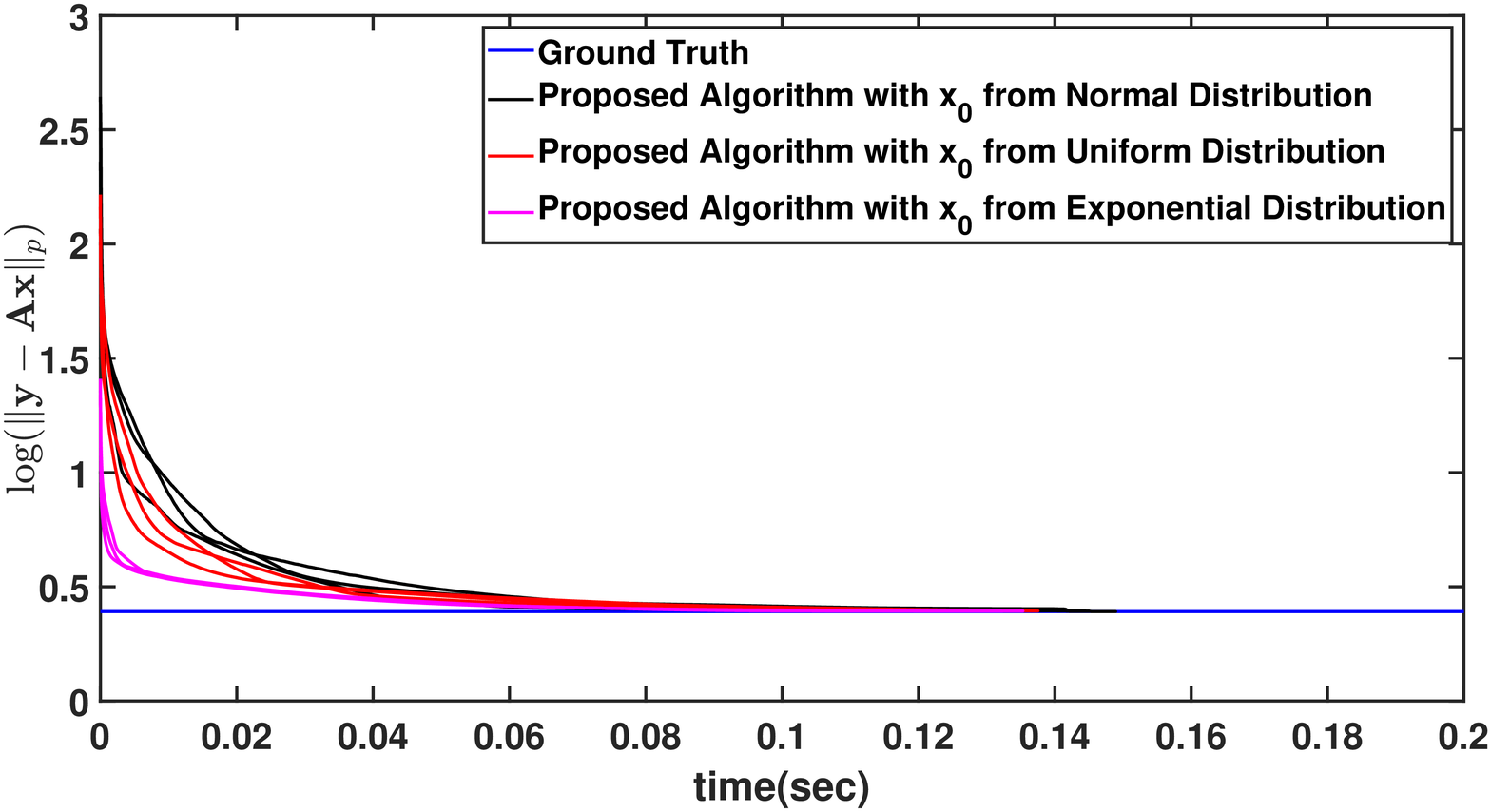}
\caption{Objective value vs run time of the proposed algorithm for $p=30$}
\end{subfigure}
\end{figure}
\begin{figure}\ContinuedFloat
    \centering
     \begin{subfigure}{0.49\textwidth}
  \includegraphics[height=2.1in,width=3.3in]{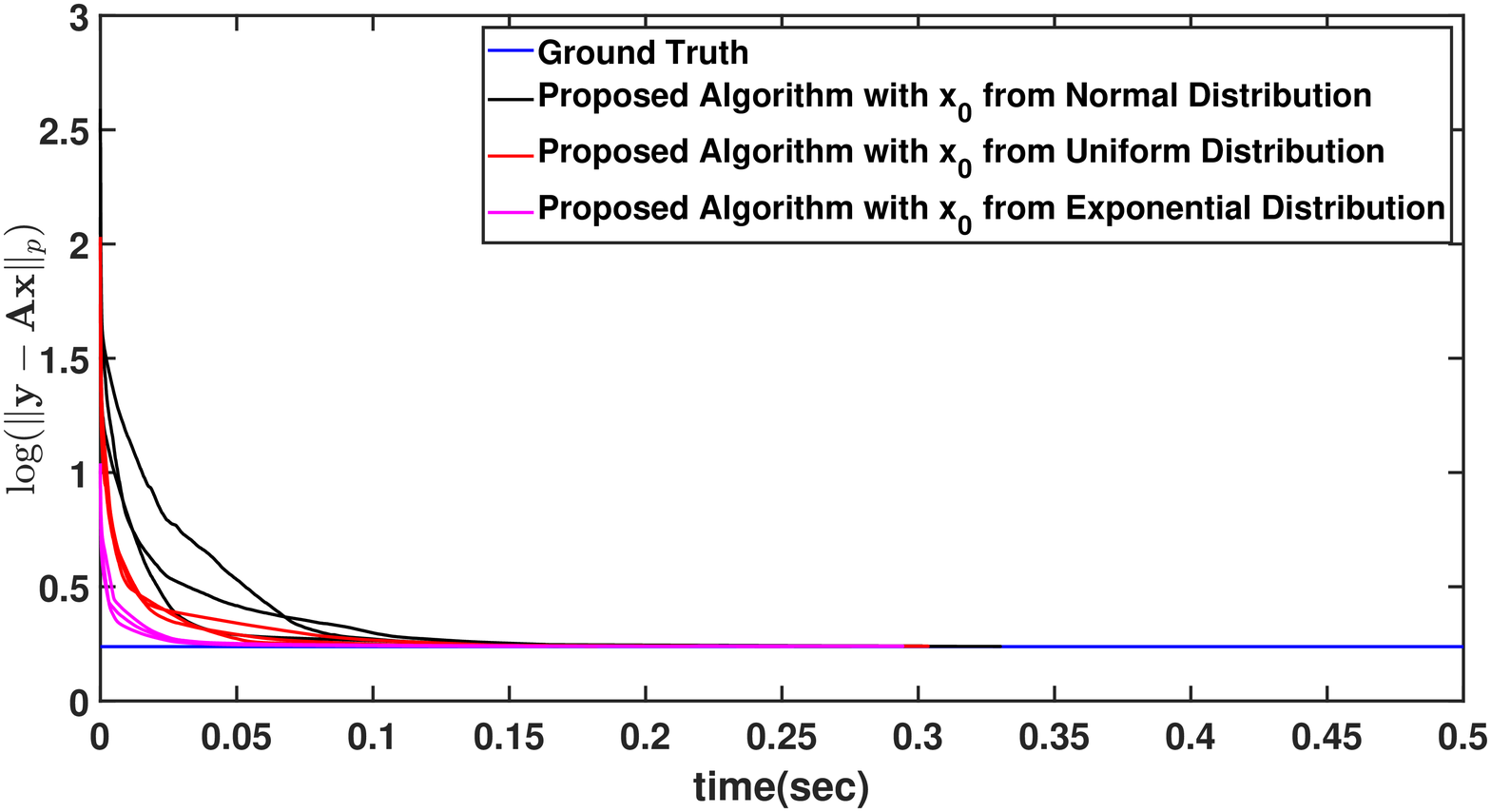}
    \caption{Objective value vs run time of the proposed algorithm for $p=80$}
    \end{subfigure}
    \label{init}
        \caption{Objective value vs run time of the proposed algorithm for different random initialization.}
\end{figure}
\\
4. In this simulation, we vary the dimension of the observation vector $\by$ and compare the convergence speed of the proposed algorithm and the modified IRLS algorithm. The dimension of the observation vector was varied i.e. $m$ was varied from $10000$ to $20000$ in steps of $1000$, $n$ was fixed at $1000$ and $p$ was kept equal to $10$. The elements of the data matrix $\bA$ and the observed data $\by$ were generated from a Normal distribution with mean zero and unit variance. Both the algorithms were initialized at the optimal point of $\ell_{2}$ norm regression problem. The run time was averaged over 50 trials. Fig. \ref{m_vary} shows the performance of the algorithms for the varying dimension of the observation vector $\by$ and as can be seen from the figure the proposed algorithm converges about five times faster than the modified IRLS algorithm. 
\begin{figure}[!h]
\centering
\begin{tabular}{c}
\includegraphics[height=2.1in,width=3.3in]{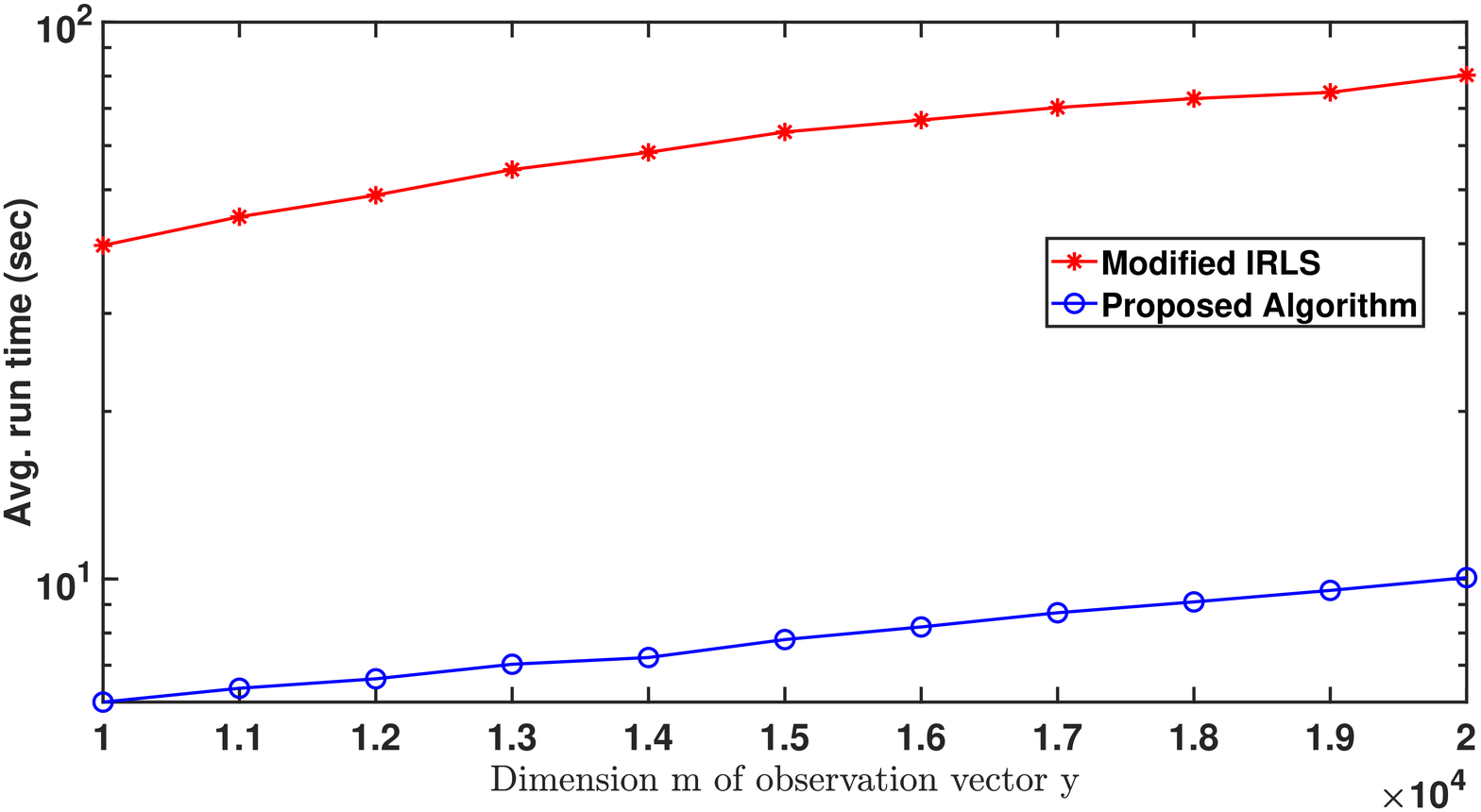}
\end{tabular}
\caption{Comparison of average run time of the proposed algorithm with the modified IRLS algorithm for varying size of the observation vector $\by$ }
\label{m_vary}
\end{figure}
\\
5. In this simulation, we show that the proposed algorithm converges to the optimal solution for the two special cases discussed in section. \ref{sec:4} i.e., for the $\ell_{1}$ and the $\ell_{\infty}$ norm linear regression problems. In the case of $\ell_{1}$ norm linear regression, we also compare the convergence speed of the proposed algorithm with the IRLS algorithm. Since, neither the IRLS algorithm nor the modified IRLS algorithm can be extended to solve the $\ell_{\infty}$ norm regression problem, we do not compare the convergence speed of the proposed algorithm with them. For both the special cases, the elements of the matrix $\bA$  and the observed data $\by$ were randomly generated from Normal distribution with zero mean and unit variance. Fig. \ref{l1} and Fig. \ref{linf} shows the objective value vs. run time of the proposed algorithm for the $\ell_{1}$ and $\ell_{\infty}$ norm regression problems for different sizes of the data matrix $A$ and for different random initialization $\bx^{0}$. From Fig. \ref{l1} and Fig. \ref{linf} it can be seen that the proposed algorithm converges to the true solution for all the initialization. Also, from Fig. \ref{l1} it can be seen that the proposed algorithm converges faster than the IRLS algorithm. 
\begin{figure}[!h]
\centering
\begin{subfigure}{0.49\textwidth}
\centering
\captionsetup{justification=centering}
\includegraphics[height=2.1in,width=3.3in]{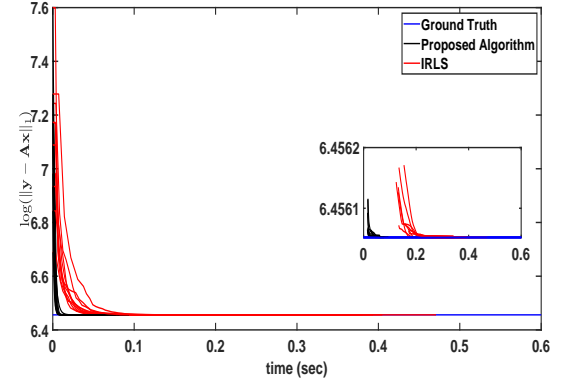}
\caption{Objective value vs run time for problem dimension $m=800$ and $n=3$}
\end{subfigure}
\begin{subfigure}{0.49\textwidth}
\centering
\captionsetup{justification=centering}
\includegraphics[height=2.1in,width=3.3in]{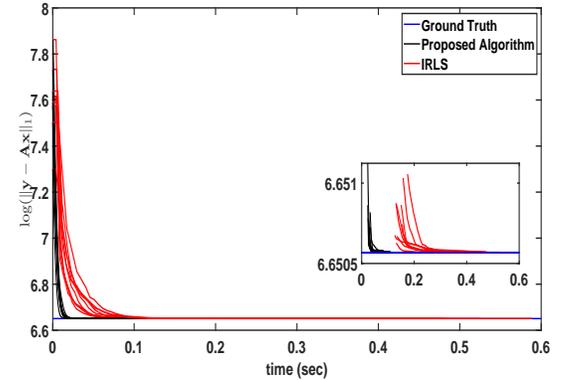}
\caption{Objective value vs run time for problem dimension $m=1000$ and $n=5$.}
\end{subfigure}
\caption{Objective value vs run time of the proposed and the IRLS algorithm for the $\ell_{1}$ norm regression problem}
\label{l1}
\end{figure}
\vspace{-1mm}
\begin{figure}[!h]
\centering
\begin{subfigure}{0.49\textwidth}
\centering
\captionsetup{justification=centering}
\includegraphics[height=2.1in,width=3.3in]{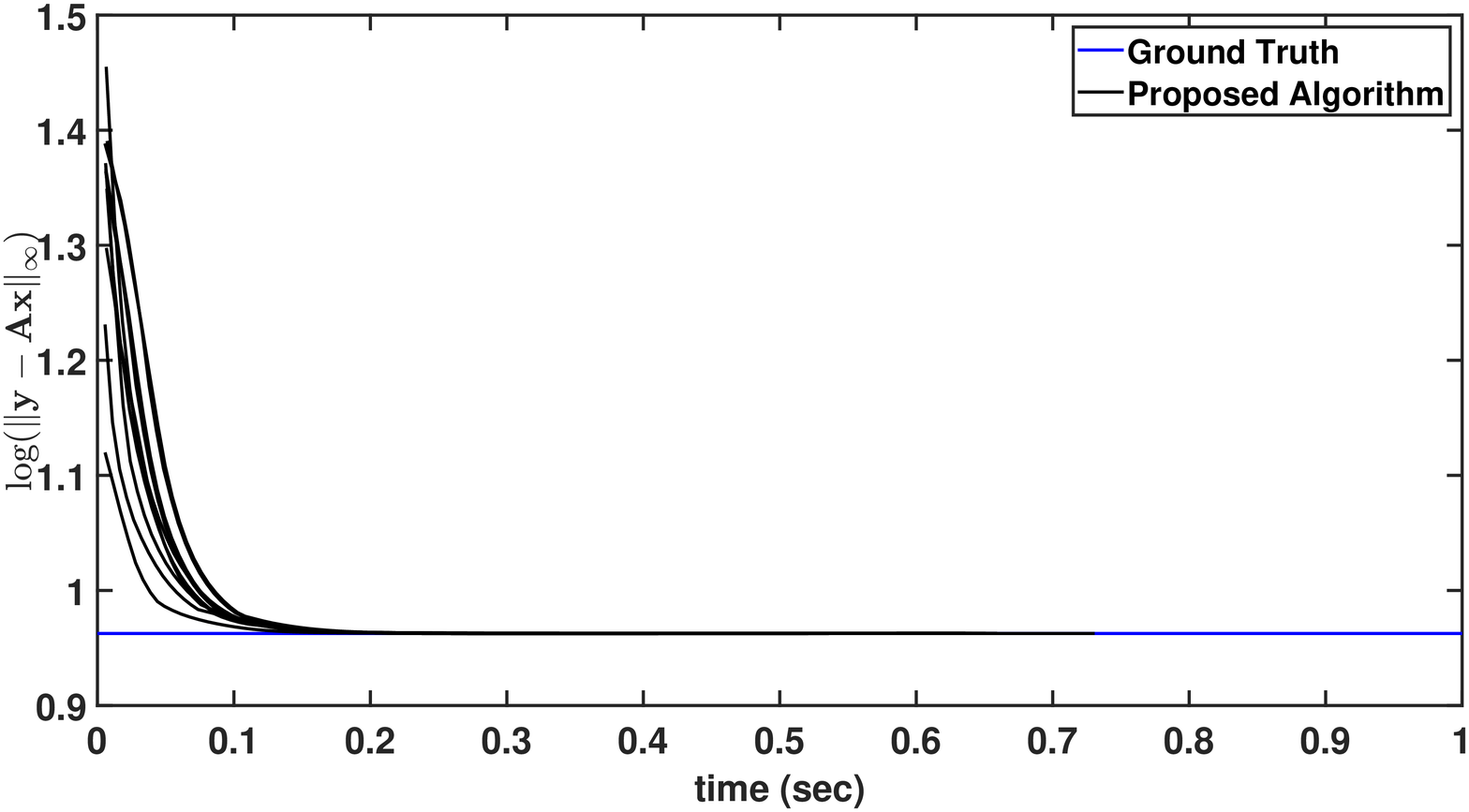}
\caption{Objective value vs run time for problem dimension $m=500$ and $n=3$}
\end{subfigure}
\begin{subfigure}{0.49\textwidth}
\centering
\captionsetup{justification=centering}
\includegraphics[height=2.1in,width=3.3in]{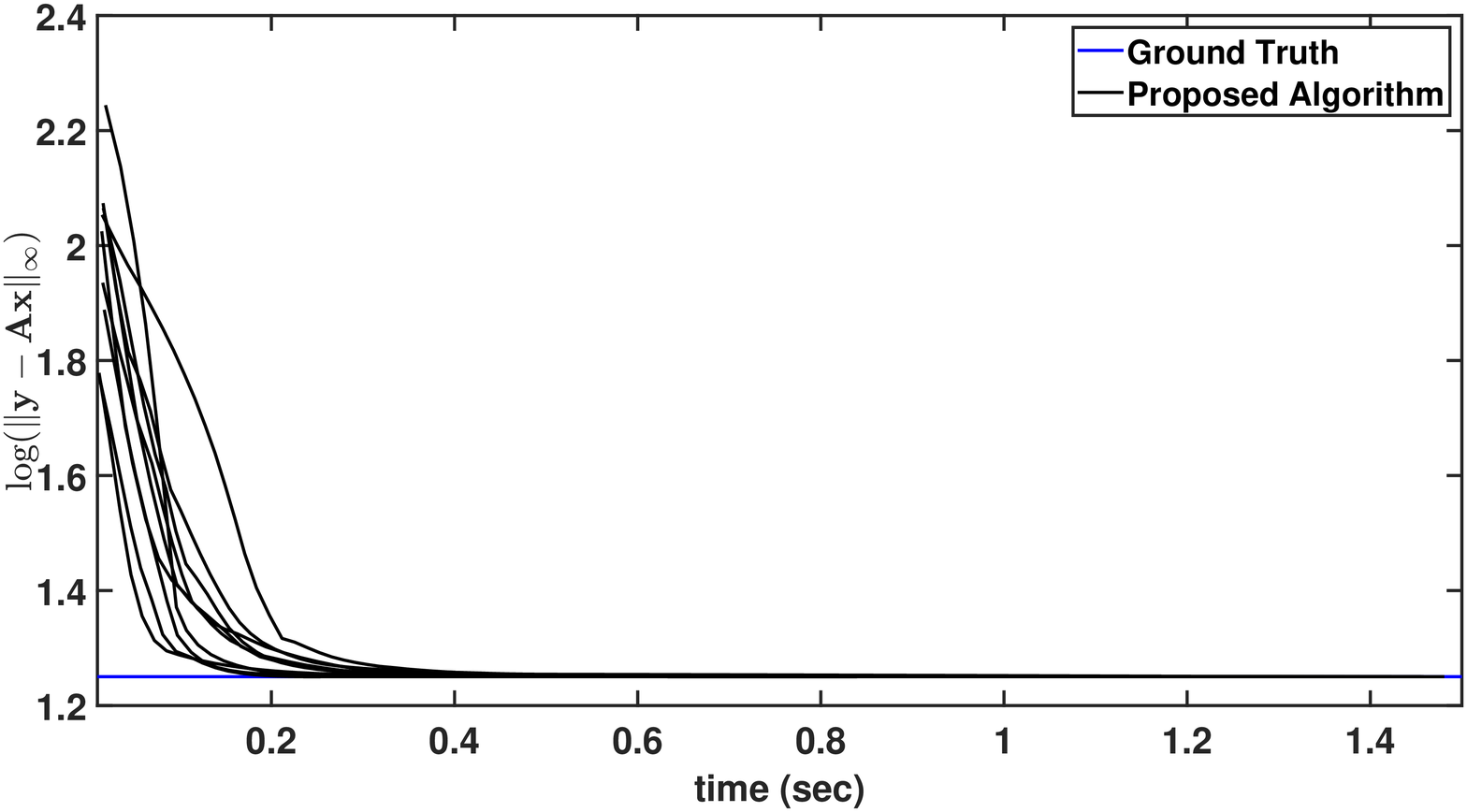}
\caption{Objective value vs run time for problem dimension $m=1000$ and $n=5$.}
\end{subfigure}
\caption{Objective value vs run time of the proposed algorithm for the $\ell_{\infty}$ norm regression problem}
\vspace{-9.61mm}
\label{linf}
\end{figure}
6. In this simulation we compare the performance of the proposed algorithm and the modified IRLS algorithm for the Graph based semi-supervised problem. To do so, we simulated the labeled and unlabeled data points randomly from a uniform distribution from $[0,1]^{d}$ for $d=10$. The labels were also generated randomly from a uniform distribution from $[0,1]$. A K-NN graph was constructed with the value of $K=10$ using the code in \cite{code}. The non-negative edge weights were generated using (\ref{weights}). We compared the convergence of the algorithms by varying the number of unlabeled data points $u$ from $50$ to $500$ in steps of $50$ and for fixed number of labeled data points equal to $10$. Fig. \ref{synthetic} compares the average run time of the algorithms for varying number of unlabeled points. The run time was average over $50$ monte carlo trials. From Fig. \ref{synthetic} it can be seen as the number of unlabeled points increases the run time of the modified IRLS algorithm increases exponentially while that that of the proposed algorithm increases linearly. 
\begin{figure}[H]
\centering
\captionsetup{justification=centering}
\includegraphics[height=2.1in,width=3.3in]{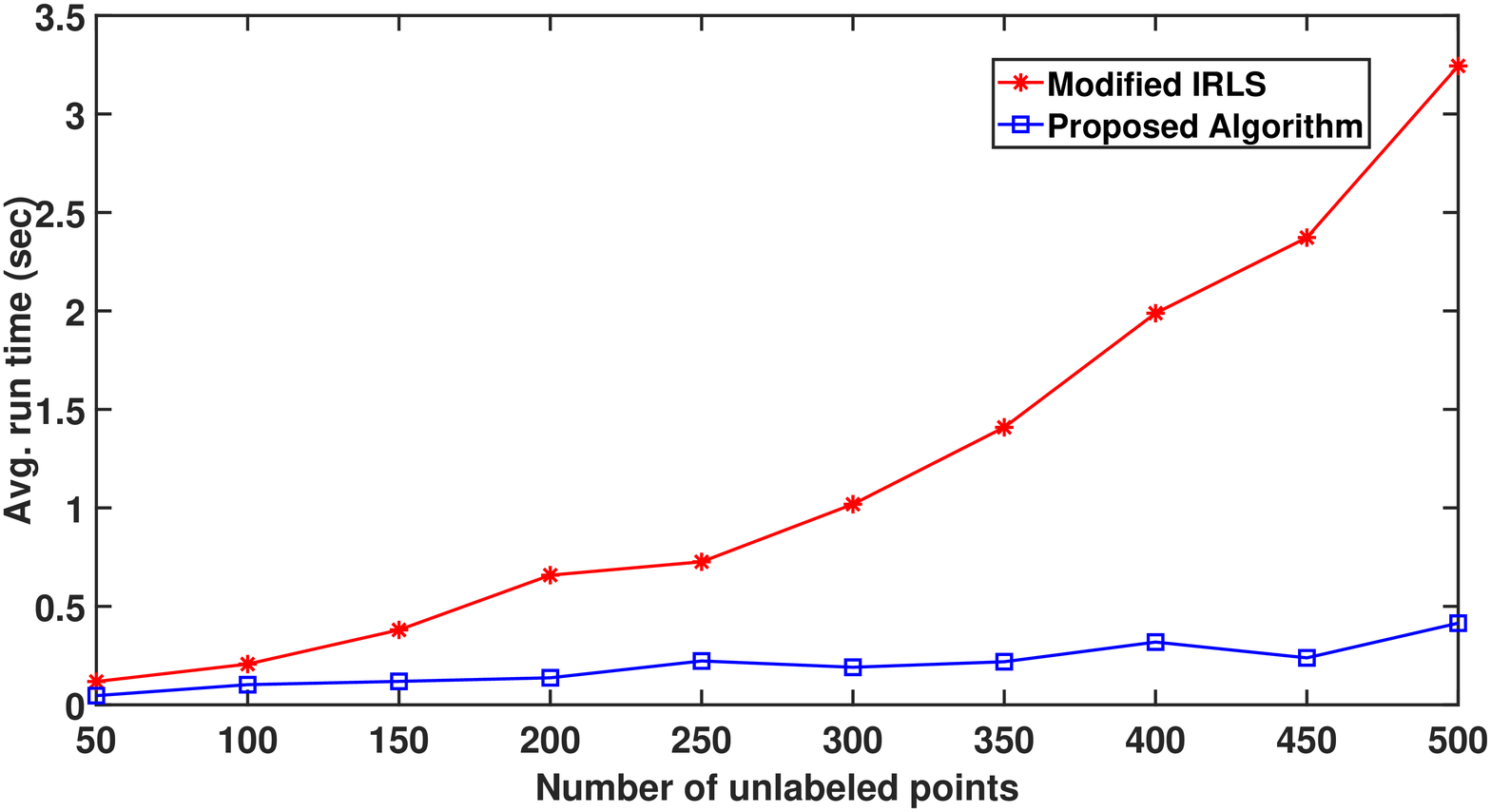}
\caption{Comparison of average run time of the proposed algorithm with the modified IRLS algorithm for simulated data.}
\label{synthetic}
\end{figure}
Next, we compare the performance of the algorithms for the graph based semi-supervised problem on two different real data sets obtained from \cite{data}.  We describe each data set here:
\begin{enumerate}
    \item{IRIS data set - This data set contains fifty samples of three different Iris flower - Iris Setosa, Iris Versicolour and Iris Virginica. Four features were measured from each sample - sepal length, sepal width, petal length and the petal width, all in centimeter. Hence, the dimension $d$ of the data points is equal to four. The task is to learn a classifier which can can classify the data points into its corresponding species.}
    \item{SPECT data set - It is a multivariate data set containing $267$ samples. Each sample is described by $23$ features obtained from the Single Proton Emission Computed Tomography (SPECT) images. Hence, the dimension $d$ of each data point is equal to $23$. The task is to classify the given data point as normal or abnormal perfusion.}
\end{enumerate}
Similar to the previous experiment we compare the performance of the algorithms by varying the number of unlabeled data points. The unlabeled data points were obtained by taking $l_{c}$ data points from each class such that the total number of labeled data points is equal to $l$ and treating the remaining data as unlabeled data. Table. \ref{iris} and Table. \ref{spect} compares the run time of the algorithms for 
varying number of labeled points per class and from the Table it can be seen that the proposed algorithm converges faster than the modified IRLS algorithm for both the data sets.

\begin{table}[!h]
\centering
\caption {Comparison of run time of algorithms in seconds for IRIS data set}
\label{iris}
\begin{tabular}{|p{2.5cm}|p{2.5cm}|p{2.1cm}|}
\hline
No. of labeled data points per class&Proposed algorithm & Modified IRLS \cite{firls}\\
 \hline
$2$ &$0.18$&$0.29$\\
$4$ &$0.14$&$0.28$\\
$6$ &$0.11$&$0.25$\\
$8$ &$0.08$&$0.24$\\
$10$&$0.07$&$0.21$\\
 \hline
\end{tabular} 
\end{table}

\begin{table}[!h]
\centering
\caption {Comparison of run time of algorithms in seconds for SPECT data set}
\label{spect}
\begin{tabular}{|p{2.5cm}|p{2.5cm}|p{2.1cm}|}
\hline
No. of labeled data points per class&Proposed algorithm & Modified IRLS \cite{firls}\\
 \hline
$2$ &$0.19$&$1.01$\\
$4$ &$0.16$&$0.97$\\
$6$ &$0.11$&$0.91$\\
$8$ &$0.09$&$0.86$\\
$10$&$0.08$&$0.78$\\
 \hline
\end{tabular} 
\end{table}

\section{Conclusion}\label{sec:6}
In this paper, we proposed an iterative algorithm \textbf{PROMPT} based on the MM procedure to solve the $\ell_{p}$ norm regression problem for any value of $p$. \textbf{PROMPT} can parallely update each element of $\bx$ which is helpful to handle the large scale data efficiently. Also, unlike the state-of-the-art algorithm, the proposed algorithm converges to to the optimal point irrespective of the initialization scheme. We show through computer simulations that the proposed algorithm has faster speed of convergence when compared to the state-of-the-art algorithms and in the end we also evaluate the performance of the proposed algorithm for the graph based semi-supervised problem using real and simulated data. 
\bibliographystyle{IEEEtran} 
\bibliography{ref}
\end{document}